\documentclass{article}
\usepackage{epsfig,graphics}
\usepackage{xcolor}
\usepackage{epstopdf}
\definecolor{mydarkblue}{rgb}{0,0.08,0.45}
\usepackage[bookmarks, colorlinks=true, citecolor=mydarkblue,linkcolor=mydarkblue,urlcolor=mydarkblue]{hyperref}
\usepackage{url}
\usepackage{times}
\usepackage{amsmath,amssymb}
\usepackage{amsthm}
\usepackage{mathrsfs}
\usepackage{mhchem}
\usepackage{subfigure}
\usepackage{natbib}
\usepackage{algorithm}
\usepackage{algorithmic}

\usepackage[accepted]{icml2017}

\newcommand{\rd}{\textrm{d}}

\newcommand{\eps}{\epsilon}

\newcommand{\vq}{\mathbf{q}}
\newcommand{\vr}{\mathbf{r}}
\newcommand{\vone}{\mathbf{1}}
\newcommand{\vpi}{\mbox{\boldmath$\pi$}}

\newcommand{\snorm}{{|\mathcal{S}|}}

\newcommand{\mD}{\mathbf{D}}
\newcommand{\mB}{\mathbf{B}}
\newcommand{\mV}{\mathbf{V}}

\newcommand{\psU}{\widetilde{U}}
\newcommand{\mQ}{\mathbf{Q}}

\newcommand{\vx}{\mathbf{x}}
\newcommand{\vy}{\mathbf{y}}

\newcommand{\rt}{{\rm T}}

\newsavebox{\measurebox}
\icmltitlerunning{Stochastic Gradient MCMC Methods for Hidden Markov Models}

\begin{document}
\twocolumn[
\icmltitle{Stochastic Gradient MCMC Methods for Hidden Markov Models}
\begin{icmlauthorlist}
\icmlauthor{Yi-An Ma}{1}
\icmlauthor{Nicholas J. Foti}{1}
\icmlauthor{Emily B. Fox}{1}
\end{icmlauthorlist}
\icmlaffiliation{1}{University of Washington, Seattle, WA, USA}
\icmlcorrespondingauthor{Yi-An Ma}{yianma@uw.edu}
\vskip 0.3in
]
\printAffiliationsAndNotice{}

\begin{abstract}
Stochastic gradient MCMC (SG-MCMC) algorithms have proven useful in scaling Bayesian inference to large datasets
%been successfully applied in Bayesian inference with large datasets
under an assumption of i.i.d\ data.
We instead develop an SG-MCMC algorithm to learn the parameters of hidden Markov models (HMMs) for time-dependent data.
There are two challenges to applying SG-MCMC in this setting: The latent discrete states, and needing to break dependencies when considering minibatches.
%The challenge in applying SG-MCMC to dependent data is the need to break the dependencies when considering minibatches of observations.
We consider a marginal likelihood representation of the HMM and propose an algorithm that harnesses the inherent memory decay of the process.
We demonstrate the effectiveness of our algorithm on synthetic experiments and an ion channel recording data, with runtimes significantly outperforming batch MCMC.
% In terms of runtime, our algorithm significantly outperforms the batch MCMC algorithm.
\end{abstract}

\section{Introduction}
\label{sec:intro}

Stochastic gradient based algorithms have proven crucial in numerous areas for
scaling inference algorithms to large datasets. The key idea is to employ noisy estimates
of the gradient based on minibatches of data, avoiding a costly gradient
computation using the full dataset \citep{SGD}. Assuming the data are i.i.d.\ and the minibatches are properly scaled,
the stochastic gradient is an unbiased estimate of the true gradient.  In the
context of Bayesian inference, such approaches have proven useful in scaling
variational inference~\citep{SVI,OnlineVB,StreamingVB,Foti:Xu:Laird:Fox:2014}
and Markov chain Monte Carlo (MCMC)~\citep{SGLD,SGRLD,SGHMC,SGNHT,Shang}.  For
the latter, a primary focus has been on the influence of the stochastic
gradient noise on the MCMC iterates; in contrast to many optimization-based
procedures, it is non-trivial to show that the underlying (stochastic) dynamics
maintain the correct stationary distribution in the presence of such noise.
Significant headway has been made in developing such correct SG-MCMC
procedures.
%For example, recently a recipe was proposed that translates the
%challenging problem of constructing efficient SG-MCMC algorithms into one of
%simply selecting two matrices~\citep{completesample}.
These algorithms have shown great practical benefits and have
gained significant traction.

A separate challenge, however, is the important and often overlooked question of whether such stochastic gradient techniques can be applied to massive amounts of \emph{sequential} or otherwise \emph{non-i.i.d.} data.  In such cases, crucial dependencies must be broken to form the necessary minibatches.  This question received some attention in the stochastic variational inference (SVI) algorithm of~\citet{Foti:Xu:Laird:Fox:2014} for hidden Markov models (HMMs).  In this work, we also focus in on HMMs as a simple example of a sequential data model, but turn our attention to SG-MCMC algorithms.

%Recently, important progress has been made in stochastic gradient MCMC (SG-MCMC) methods to deal with the task of Bayesian inference for large data sets \cite{SGLD,SGRLD,SGHMC,SGNHT,Shang}.
%A complete recipe has been proposed so that the task of constructing efficient SG-MCMC algorithms is transferred to the choice of two matrices \cite{completesample}.
%The key idea is to employ noisy estimates of the gradient based on minibatches of data, avoiding a costly gradient computation using the full dataset \cite{SGD}.
%Assuming that the data are identical independently distributed, the stochastic gradient is equivalent to the true gradient over the entire dataset plus a Gaussian random noise.
%One question is often asked with these stochastic gradient techniques: can they be applied to problems with massive amounts of sequential data? In this work we develop an SG-MCMC approach to efficiently sample from the posterior over model parameters for sequential data modeled with a hidden Markov model (HMM).

There are many existing algorithms for inferring the model parameters of an HMM including Monte Carlo methods~\citep{Scott:2002}, expectation-maximization (EM)~\citep{Bishop}, and variational algorithms~\citep{Beale:2003}. All of these ideas operate by iterating between a \textit{local update} for the latent states, followed by a \textit{global update} of the model parameters. The local update is usually performed using the \textit{forward-backward} algorithm that allows computation of any marginal, or pair-wise marginal, in time linear in the length of the sequence.

In the variational context, recent work has focused on scaling these \textit{local-global} inference schemes to settings with a large number of replicates of short sequences~\citep{Johnson:Willsky:2014,Hughes:Stephenson:Sudderth:2015}. These methods utilize the fact that independent replicates of the observation sequence can be used to compute unbiased gradient estimates~\citep{Johnson:Willsky:2014}, and can be used to incrementally update sufficient statistics~\citep{Hughes:Stephenson:Sudderth:2015}. In contrast, the SVI-HMM algorithm of~\citet{Foti:Xu:Laird:Fox:2014} examines how to deal with extremely long observation sequences. The algorithm heuristically breaks the dependence between observations and performs local updates on short subsequences of observations using a limited forward-backward algorithm. These existing methods suffer from a number of drawbacks. The variational approaches must use an approximate posterior distribution for both the state- and model-parameters, which may not be representative of the true distributions. The methods are also limited to conjugate prior distributions over the parameters, which can severely limit the expressiveness of the model.
Finally, all of the methods discussed thus far are susceptible to the widely known problem of underestimating posterior correlations biasing fully Bayesian analyses.

Unfortunately, attempting to naively use subchains as in \citet{Foti:Xu:Laird:Fox:2014} within SG-MCMC approaches is fraught with difficulty: The local-global structure of SVI-HMM does not lend itself to deriving provably correct SG-MCMC algorithms, stemming from two main challenges.

%Unfortunately, existing SG-MCMC approaches cannot be adapted to HMMs by simply deploying the ideas of~\citet{Foti:Xu:Laird:Fox:2014} within the MCMC context.
The first challenge is that SG-MCMC methods sample continuous-valued parameter representations, whereas the HMM learning objective is typically specified in terms of the discrete-valued state sequence (local variables).  To address this challenge, we consider an alternative approach to performing parameter inference for HMMs: We work directly with the marginal likelihood of the observation. We form stochastic gradients by only evaluating terms of the full gradient that depend on a small subsequence.
%Our stochastic gradients are then formed by \textit{evaluating only subsequence-specific terms of the gradient of the marginal likelihood}. %\textit{evaluate the gradient of the marginal likelihood on small subsequences of observations}.

The second challenge is handling the temporal dependencies, specifically: i) each subsequence-specific term in the stochastic gradient still requires a forward-backward pass on the rest of the sequence, and ii) proximal subsequences are mutually correlated. We address both of these issues by capitalizing on the well-known memory decay of the Markovian structure underlying the data generating process. Specifically, we approximate the full forward-backward passes with message passing only on short buffers around the considered subsequences of observations.  We further restrict subsequences to be sufficiently far from one another to ensure that computations with them are uncorrelated.  We provide a \textit{theoretically justified} approach to estimating this buffer length and between-subsequence gap, allowing us to prove the validity of the resulting SG-MCMC algorithm.  In particular, the same theoretical guarantees are provided as in the i.i.d. setting.

Buffering to perform limited message passing in HMMs was also applied in SVI-HMM~\citet{Foti:Xu:Laird:Fox:2014}; however, the buffering there was part of a latent state update. In particular, SVI-HMM iterates between buffered message passing for local updates and stochastic gradients for global updates.  We, in contrast, consider stochastic gradients of a marginal likelihood representation and utilize buffering directly within this stochastic gradient calculation.

We evaluate the efficacy of our buffered SG-MCMC method for HMMs on two synthetic examples with very different dynamics. We compare against an unbuffered SG-MCMC approach as well as against treating the data as i.i.d. Finally, we show the computational gains of our SG-MCMC algorithm over batch MCMC by segmenting an ion channel dataset where a 1,000X speedup was observed. Collectively, our contributions make a sizable step towards general purpose SG-MCMC algorithms for sequential data.

%\nf{We will probably need to update this when we have the final synthetic experiments.}
%Our synthetic data experiments investigate the impact of errors introduced via treating the data as i.i.d. or
%considering subchains naively, and how our buffering scheme can alleviate these
%issues. We then explore the benefits of our method over SVI-HMM.
%Finally, we provide an exploration of the computational gains over batch MCMC
%methods in an ion channel segmentation task. Here, our SG-MCMC algorithm
%provides good performance in segmenting the ion channel data about $1,000$ times faster than the batch MCMC method.
%%We further demonstrate the benefits of the stochastic gradient in an ion channel data segmentation experiment.
%%Our SG-MCMC algorithm provides a good segmentation within \textbf{Yian: [insert time here]} whereas a single iteration of batch MCMC takes \textbf{Yian: [insert time here]}.
%Collectively, our contributions make a sizable step towards general purpose SG-MCMC algorithms for sequential data.

\section{Background}
\subsection{Hidden Markov Models}
\label{sec:HMM}
Hidden Markov models (HMMs) are a class of discrete-time doubly stochastic processes consisting of a i) latent discrete-valued state sequence $x_t \in \{1, \ldots, K\}$ generated by a Markov chain and (ii) corresponding observations $y_t$ generated from distributions determined by the latent states $x_t$.
Specifically, the joint distribution of $\vy = (y_1,\cdots,y_T)$ and $\vx = (x_0,\cdots,x_T)$, factorizes as
\begin{align}
p(\vx,\vy) = \pi_0(x_0) \prod_{t=1}^T p(x_t|x_{t-1},A) \cdot p(y_t|x_t,\phi),
\end{align}
where $A$ is the Markov transition matrix such that $A_{i,j} = \Pr(x_t=i | x_{t-1}=j)$, $\{\phi_k\}^K_{k=1}$ are the emission parameters, and $\pi_0 = p(x_0)$ is the initial state distribution.
We denote the parameters of interest as $\theta = \{A, \phi\}$ and do not focus on performing inference on $\pi_0$.

Traditionally, EM, variational inference, or MCMC are used to perform inference over $\theta$~\citep{Scott:2002,Beale:2003}. These algorithms rely on the well-known \textit{forward-backward algorithm} to compute the marginal, $p(x_t|y_{1:T})$, and pairwise-marginal, $p(x_t,x_{t+1}|y_{1:T})$, distributions. The algorithm works by recursively computing a sequence of forward messages $\alpha_t(x_t) = p(x_t|y_{1:t})$ and backwards messages $\beta_t(x_t) = p(y_{t+1:T}|x_t)$ which can then be used to compute the necessary marginals~\citep{Beale:2003}. These marginals are then used to update or sample from the distribution of the model parameters.

These past algorithms have found widespread use in statistics and machine learning. However, as discussed in Sec.~\ref{sec:intro}, an alternative formulation of the HMM can provide greater utility in developing an SG-MCMC approach.
Marginalizing over $\vx$, we obtain the \textit{marginal likelihood}:
\begin{align}
\label{eq:hmm_ml}
p(\vy|\theta) = \vone^\rt \ P(y_T) A \cdots P(y_1) A \ \vpi_0,
\end{align}
where $P(y_t)$ is a diagonal matrix with $P_{i,i}(y_t) = p(y_t | x_t=i, \phi_i)$;
$\vone^\rt$ is a row vector of $K$ ones;
and $(\vpi_0)_i = \pi_0(x_0 = i)$.
The resulting \textit{posterior distribution} of $\theta$ given $\vy = y_{1:T}$ is then:
\begin{align}
\label{eq:hmm_post}
p(\theta | \vy) &\propto %p(\vy|\theta) p(\theta).
\vone^\rt \ P(y_T) A \cdots P(y_1) A \ \vpi_0 \cdot p(\theta).
%&= \vq_T^\rt \ P(y_T) A \cdots P(y_1) A \ \vpi_0 \cdot p(\theta).
\end{align}
Working with the marginal likelihood and posterior alleviates the need to
compute the marginals and pairwise marginals of $x_t$. As such, only the
forward pass of the forward-backward algorithm is performed. Indeed, performing
the matrix multiplications in Eq.~\eqref{eq:hmm_ml} from right to left
corresponds to computing the normalizing constants of the forward messages. 
Performing the matrix multiplies from left to right corresponds to
unnormalized messages in belief propagation, \citep[cf.][]{Fox:2009}.
Perhaps most importantly for the development of our SG-MCMC algorithm, the marginal likelihood does not involve alternately updating the local state variables, $x_t$, and the global model parameters $\theta$. Instead, we need only explore a continuous space which will allow us to leverage gradient information to develop a computationally and statistically efficient algorithm. The major impediment to directly using Eq.~\eqref{eq:hmm_ml} for SG-MCMC is that it is unclear how to form a stochastic gradient based on a subsequence to avoid the computational burden of gradient computations in large $T$ settings.

\subsection{Stochastic Gradient MCMC for i.i.d. Data}
\label{sec:SGMCMCiid}
One approach for devising MCMC algorithms is to utilize continuous dynamics to explore a potential function $U(\theta) \propto -\ln \pi(\theta)$ for target distribution $\pi(\theta)$; for Bayesian inference, we take $U(\theta) \propto -\ln p(\theta|\vy)$, i.e., the negative log posterior. Then, samples of a continuous valued parameter, $\theta \in \mathbb{R}^d$, can be drawn as~\cite{completesample,complete_new}
\begin{align}
\theta^{(t+1)} \leftarrow& \theta^{(t)} -  \eps_{t} \left[\big({\mD}(\theta^{(t)}) + {\mQ}(\theta^{(t)})\big) \nabla{U}(\theta^{(t)}) + \Gamma(\theta^{(t)})\right] \nonumber\\
&+ \mathcal{N}(0, \eps_{t} (2\mD(\theta^{(t)}))),  \label{eq:update}
\end{align}
where $\Gamma_i(\theta) = \sum_{j=1}^d \dfrac{\partial}{\partial \theta_j} (\mD_{i,j}(\theta) + \mQ_{i,j}(\theta))$, ${\mD}(\theta^{(t)})$ is a positive-definite matrix and ${\mQ}(\theta^{(t)})$ a skew-symmetric matrix. \citet{completesample} proved that in the limit $\epsilon_t \rightarrow 0$ with ergodicity, the iterates $\theta^{(t)}$ will be drawn from $p(\theta|\vy)$. % \footnote{In the occasions when Euler-Maruyama integration scheme shown in Eq.~\ref{eq:update} is not stable, a more stable integration of the dynamics of Eq.~\ref{eq:update} is required.}.

%We leverage the SG-MCMC framework of~\citet{completesample,complete_new} which found that any sampler using continuous Markov dynamics can be represented as follows.
%If we define the potential function $U(\theta) \propto -\ln p(\theta|\vy)$ as the negative log posterior,
%then any such sampler that preserves the posterior distribution has an update rule:
%\begin{align}
%\theta^{(t+1)} \leftarrow& \theta^{(t)} -  \eps_{t} \left[\big({\mD}(\theta^{(t)}) + {\mQ}(\theta^{(t)})\big) \nabla{U}(\theta^{(t)}) + \Gamma(\theta^{(t)})\right] \nonumber\\
%&+ \mathcal{N}(0, \eps_{t} (2\mD(\theta^{(t)}))),  \label{eq:update}
%\end{align}
%where $\Gamma_i(\theta) = \sum_{j=1}^d \dfrac{\partial}{\partial \theta_j} (\mD_{i,j}(\theta) + \mQ_{i,j}(\theta))$, ${\mD}(\theta^{(t)})$ is any positive-definite matrix and ${\mQ}(\theta^{(t)})$ is any skew-symmetric matrix.

For i.i.d.\ data, ${U}(\theta) = -\sum_{{s}\in {\mathcal{S}}} \ln p({y_s}|\theta) - \log p(\theta)$.
For \emph{independently sampled} data
subsets, $\widetilde{\mathcal{S}} \subset \mathcal{S}$, a noisy unbiased estimate of
the potential function is given by:
\begin{align}
\widetilde{U}(\theta) = -\dfrac{|\mathcal{S}|}{|\widetilde{\mathcal{S}}|} \sum_{s\in \widetilde{\mathcal{S}}} \log p({y_s}|\theta) - \log p(\theta); \quad \widetilde{\mathcal{S}} \subset \mathcal{S}.
\label{eqn:tildeU}
\end{align}
As such, a gradient computed based on $\widetilde{U}(\theta)$---called a \emph{stochastic gradient}%~\cite{SGD}
---is a noisy, but unbiased estimator of the full-data gradient.  The key question %in many of the existing SG-MCMC algorithms 
is whether the noise injected by the stochastic gradient adversely affects the stationary distribution of the modified dynamics (using $\nabla\widetilde{U}(\theta)$ in place of $\nabla U(\theta)$). One way to analyze the impact of the stochastic gradient is to make use of the central limit theorem and assume
\begin{equation}
\nabla \psU(\theta) = \nabla U(\theta) +\mathcal{N}(0, \mV(\theta)). \label{eqn:noisygrad}
\end{equation}
Simply using $\nabla \psU(\theta)$ in place of $\nabla U(\theta)$ in Eq.~\eqref{eq:update} results in an additional noise term $({\mD}(\theta) + {\mQ}(\theta)\big)\ \mathcal{N}(0,\mV(\theta))^T$.
Assuming we have an estimate $\widehat{\mB}$ of the variance of this additional noise satisfying $2\mD(\theta) - \eps \widehat{\mB} \succeq 0$ (i.e., positive semidefinite),
%With small $\epsilon$, this is always true since the stochastic gradient noise %$\eps \mV(\theta)$
%scales down faster than the added noise.
then we can attempt to account for the stochastic gradient noise by simulating
\begin{align}
%\begin{tiny}
\theta^{(t+1)} \leftarrow& \theta^{(t)} -  \eps_{t} \left[\big({\mD}(\theta^{(t)}) + {\mQ}(\theta^{(t)})\big) \nabla \widetilde{U}(\theta^{(t)}) + \Gamma(\theta^{(t)})\right] \nonumber\\
\label{eq:true-update}
&+ \mathcal{N}(0, \eps_{t} (2\mD(\theta^{(t)}) - \eps_{t} \widehat{\mB}^{(t)})).
%\end{tiny}
%\label{eq:sgmcmc_update}
\end{align}
This is the %\emph{stochastic gradient} MCMC (SG-MCMC) 
SG-MCMC algorithm for i.i.d. data proposed by~\citet{completesample,complete_new}. See Alg. \ref{alg:SG_MCMC}.

For this SG-MCMC, there are sources of error introduced via (i) discretizing the continuous stochastic dynamics and (ii) estimation of the stochastic gradient noise covariance. Although the algorithm is provably correct as $\epsilon_t \rightarrow 0$, in practice one uses a small, finite stepsize for greater efficiency. In such cases, bias is introduced. This bias-variance tradeoff was recently studied in~\cite{Vollmer}. Higher order numerical schemes \cite{DingHighOrder,Shang_new} and a moving window estimation of $\widehat{B}$ can further reduce this bias \cite{Shang}.

\begin{algorithm}[t!]
\caption{SG-MCMC}
\begin{algorithmic}
\STATE initialize $\theta_0$
 \FOR{$t=0,1,2\cdots,N_{iter}$}
    \FOR{$i=1\cdots n$}
    \STATE
    \begin{small}
    $\Gamma_i(\theta) = \sum_j \dfrac{\partial}{\partial \theta_j} \left({\mD}_{ij}(\theta) + {\mQ}_{ij}(\theta)\right)$
    \end{small}
    \ENDFOR
    \STATE sample
    \begin{small}
    $\eta^{(t)} \sim \mathcal{N}(0, 2 \eps_{t} \mD(\theta^{(t)}) - \eps_{t}^2 \widehat{\mB}^{(t)})$
    \end{small}
   \STATE
   \begin{small}
%   $\begin{array}{cl}
%   \theta^{(t+1)} \leftarrow& \theta^{(t)} -  \eps_{t} \left[\big({\mD}(\theta^{(t)}) + {\mQ}(\theta^{(t)})\big) \nabla \widetilde{U}(\theta^{(t)})
% + \Gamma(\theta^{(t)})\right] \\
% &+ \eta^{(t)}
%   \end{array}$
   $\begin{array}{cl}
   \theta^{(t+1)} \leftarrow& \theta^{(t)} + \eta^{(t)}
   \\&-  \eps_{t} \left[\big({\mD}(\theta^{(t)}) + {\mQ}(\theta^{(t)})\big) \nabla \widetilde{U}(\theta^{(t)})
 + \Gamma(\theta^{(t)})\right]
   \end{array}$
   \end{small}
   \ENDFOR
\end{algorithmic}
\label{alg:SG_MCMC}
\end{algorithm}

\section{Stochastic Gradient MCMC for HMMs}
\label{sec:sgmcmc}
\begin{figure*}
	\includegraphics[width=\textwidth]{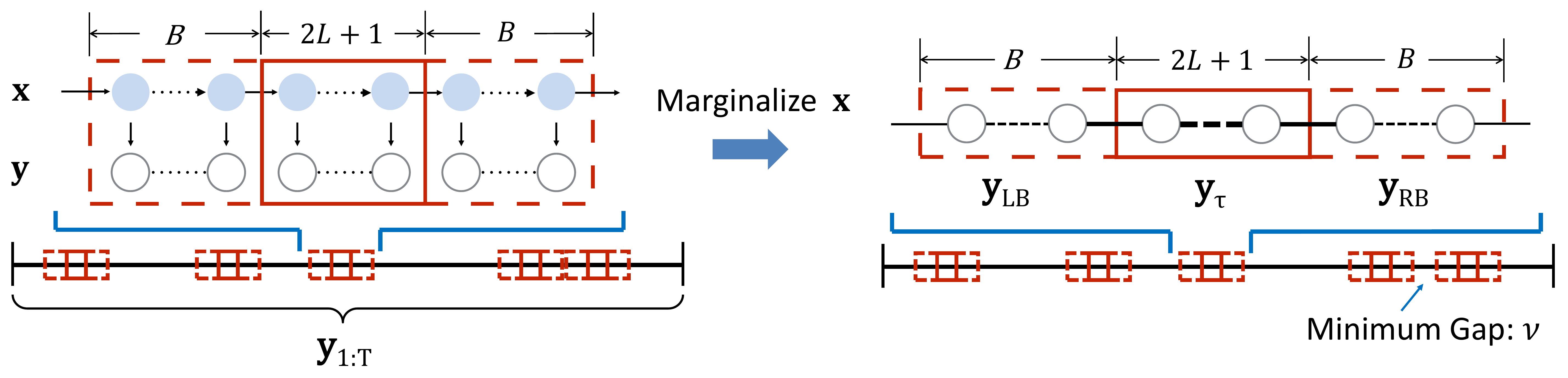}
    \vspace{-20pt}
	\caption{Diagram of subsequences, buffers, and subsequence sampling from full observation sequence.
{\em Left:} The SVI method of \citet{Foti:Xu:Laird:Fox:2014} approximates stochastic gradients using subchains of length $2L+1$ using the forward-backward algorithm performed on both the subchains and the associated buffer chains of length $B$.
{\em Right:} Our propsoed SG-MCMC method uses a similar subsampling approach,
    however, i) the latent chain is never instantiated and ii) a minimum gap
    between consecutive subchains, $\vy_{\tau,L}$, is used to ensure nearly
    uncorrelated subsequences.
The thick black lines through the observables $\vy$ represent all pairwise
    correlations between observations due to marginalization of $\vx$.
    Correlation decays with distance enabling the segmentation of the of the
    chain into subsequences.}
%that after marginalization, there is longer range correlation through $\vy$.
%But correlation decays over longer range, enabling the segmentation of the whole sequence into subchains.
\label{fig:cartoon}
\end{figure*}

In order to apply SG-MCMC methods to HMMs we must be able to efficiently estimate the gradient of the potential, $U(\theta) \propto -\ln p(\theta|\vy)$. The approach we take consists of three steps (see Fig.~\ref{fig:cartoon}). First, we marginalize out the discrete state sequence and use the marginal likelihood of the data. Next, we derive an expression for the gradient of the marginal likelihood that factorizes over disjoint subsequences. Finally, we compute an unbiased noisy estimate of the gradient by randomly sampling subsequences and show that using this estimate results in an SG-MCMC algorithm that admits the desired stationary distribution under the same conditions as in the i.i.d. case (see Sec.~\ref{sec:SGMCMCiid}).
%\nf{Is this statement too much? Didn't a reviewer want us to base this on some conditions?}.

One could have imagined an alternative approach---as in SVI-HMM---of first sampling subsequences; we could then compute an approximation of the marginal likelihood on this subsequence and treat its gradient as our stochastic gradient.  However, without the marginal likelihood information in the first place, it is not obvious how subsequences correlate with each other and consequently how to control the error resulting from subsampling.  %We also emphasize that by taking an MCMC approach, we enable full Bayesian analysis and an ability to handle non-conjugate models.
%\ebf{not sure where this sentence fits, but we want to re-emphasize this.}

%Alternatively, we could naively apply the approach of SVI-HMM that first subsampled the data, then estimated a noisy gradient based on an stochastic approximation to the objective. Our approach has the benefit of working exclusively with continuous valued parameters reaping the computational and statistical benefits of SG-MCMC algorithms, including the added ability to handle non-conjugate models.

%In order to apply the SG-MCMC methodology to HMMs we must address four problems.
%First, SG-MCMC samples a continuous space.
%Hence we explicitly marginalize out the discrete state sequence and express the likelihood $p(\vy | \theta)$ as matrix multiplication as in Eq.~\ref{eq:hmm_ml}.
%Second, we must be able to compute the gradient of the marginal likelihood efficiently for large data sets.
%Instead of using minibatches of individual observations as in standard SG-MCMC, we take the minibatches to be subsequences of consecutive observations.
%The third problem is mitigating the error introduced by breaking dependencies at the endpoints of the subsequences.
%Finally, we show that SG-MCMC admits the desired stationary distribution when using subsequences of observations.

\subsection{Gradient of Marginal Likelihood Representation}
Recall that the posterior under an HMM is given by Eq.~\eqref{eq:hmm_post}
%Eqs.~\eqref{eq:hmm_ml} and \eqref{eq:bayesthm} as:
%\begin{align}
%\label{eq:hmm_post}
%p(\theta | \vy) %&\propto p(\vy|\theta) p(\theta)
%%\nonumber\\
%&\propto \vone^\rt \ P(y_T) A \cdots P(y_1) A \ \vpi_0 \cdot p(\theta),
%\end{align}
and that the \textit{potential function} $U(\theta) \propto - \ln p(\theta|\vy)$.
%Both evaluating and computing gradients of Eq.~\eqref{eq:hmm_post} is computationally expensive when $T$ is massive, for instance in problems arising in genomics.
As will prove useful in our SG-MCMC algorithm, we rewrite the posterior in terms of a
\textit{subsequence} $\vy_{\tau,L} = (y_{\tau-L},\ldots,y_\tau,\ldots,y_{\tau+L})$ with \textit{half-width} $L$ centered at time $\tau \in \{L+1,\ldots,T-L-1\}$. The overall subsequence length is $2L+1$.  Defining
\begin{equation}
\label{eq:hmm_subchain_lik}
    P(\vy_{\tau,L}) = P(y_{\tau+L})A\cdots P(y_{\tau-L})A,
\end{equation}
%In fact, given a partition of $y_{1:T}$ into non-overlapping subchains,
%$\mathcal{S} = \{\vy_{\tau,L}\}$, we rewrite Eq.~\eqref{eq:hmm_post} as
we can rewrite Eq.~\eqref{eq:hmm_post} as
\begin{align}
    p(\theta|\vy) &\propto \vq_{\tau+L+1}^\rt P(\vy_{\tau,L}) \vpi_{\tau-L-1} \cdot p(\theta).
	\label{eq:hmm_post_short}
\end{align}
Here, $\vq_{\tau+L+1,i} = p(\vy_{\tau+L+1:T}|x_{\tau+L}=i)$ is the likelihood of the observations after $\vy_{\tau,L}$ given the value of the latent state at $\tau$, and $\vpi_{\tau-L-1,i} = p(x_{\tau-L}=i|y_{1:\tau-L-1})$ is the \textit{predictive distribution} of the latent state at $\tau$ given the observations before $\vy_{\tau,L}$. Note, we do not actually need to instantiate the latent state variables $x_{\tau-L}$ and $x_{\tau+L+1}$ as $\vq_{\tau+L+1}$ and $\vpi_{\tau-L-1}$ can be computed (in theory) via the forward-backward algorithm~\cite{Rabiner:1989,Scott:2002}.
%If we partition $y_{1:T}$ into a set of contiguous and non-overlapping subsequences of length $L$, $\Pi = \{\vy_\tau : \tau = Lk+1, k=1,3,\ldots,T-L\}$, then we can rewrite Eq.~\eqref{eq:hmm_marg_lik} as $\vone^\rt P(\vy_\tau) \cdot \vpi_0$.

Let $\mathcal{S} = \{\vy_{\tau,L}\}$ be a set of non-overlapping subsequences that cover $\vy$. The gradient of %the log-posterior from
Eq.~\eqref{eq:hmm_post_short} can be written as
\begin{align}
\label{eq:gradU_full}
&\nabla U(\theta)_i = -\dfrac{\partial \ln p(\vy | \theta)}{\partial \theta_i} - \dfrac{\partial \ln p(\theta)}{\partial \theta_i}  \\ \nonumber
    &= -\sum_{\vy_{\tau,L}\in \mathcal{S}}
    \dfrac{\vq_{\tau+L+1}^\rt \dfrac{\partial P(\vy_{\tau,L})}{\partial \theta_i} \vpi_{\tau-L-1}}{\vq_{\tau+L+1}^\rt P(\vy_{\tau,L}) \vpi_{\tau-L-1}}
- \dfrac{\partial \ln p(\theta)}{\partial \theta_i},
\end{align}
%\vspace{-0.1in}
where the equality follows from the product rule (see the Supplement for complete derivation).  Importantly, note that the gradient involves a sum over terms corresponding to \emph{all} non-overlapping subsequences of length $2L+1$.

We could imagine using $\nabla U(\theta)$ from Eq.~\eqref{eq:gradU_full} in the update rule of Eq.~\eqref{eq:update} to generate sample values of $\theta$. However, Eq.~\eqref{eq:gradU_full} is extremely computationally intensive for two reasons. First, calculating $\vq_{\tau+L+1}$ and $\vpi_{\tau-L-1}$ involves the whole sequence of length $T$. Second, one must compute $\vq_{\tau+L+1}$, $P(\vy_{\tau,L})$, and $\vpi_{\tau-L-1}$ for each $\vy_{\tau,L} \in \mathcal{S}$ in the sum; this involves $T/L$ terms, thus requiring $O(T^2)$ computation time to compute the gradient.
%expensive to compute as $\vq_{\tau+L+1}$ and $\vpi_{\tau-L-1}$ require touching nearly all $T$ observations. This is prohibitive when $T$ is massive. We instead propose to compute noisy estimates of Eq.~\eqref{eq:gradU_full} using only individual or small collections of subsequences, akin to the stochastic gradient updates of Eqs.~\eqref{eqn:tildeU}-\eqref{eq:true-update}, but for our non-i.i.d.\ scenario.

%\mya{Change to:}
%\paragraph{Computational challenges}
%of Eq.~\eqref{eq:gradU_full} is two fold:
%\begin{itemize}
%\item
%Calculation of $\vq_{\tau+L+1}$, $P(\vy_{\tau,L})$, and $\vpi_{\tau-L-1}$ involves the whole chain of length $T$;
%\item
%Summation over all ${\vy_{\tau,L}\in \mathcal{S}}$ involves $T/L$ terms.
%Hence computation budget of the gradient of the log-posterior is on the order of $T^2$.
%\end{itemize}
%This is prohibitive when $T$ is massive. We instead propose to compute noisy estimates of Eq.~\eqref{eq:gradU_full} using only individual or small collections of subsequences, akin to the stochastic gradient updates of Eqs.~\eqref{eqn:tildeU}-\eqref{eq:true-update}, but for our non-i.i.d.\ scenario.

\subsection{Stochastic Gradient Calculation}

In place of $\nabla U(\theta)$ in Eq.~\eqref{eq:gradU_full}, we can define a stochastic gradient based on a single subsequence, $\widetilde{\nabla U}(\theta)$:
\begin{align}
\label{eq:gradU_single}
    \widetilde{\nabla U}(\theta)_i =
	 -\dfrac{\vq_{\tau+L+1}^\rt \dfrac{\partial P(\vy_{\tau,L})}{\partial \theta_i} \vpi_{\tau-L-1}}{\vq_{\tau+L+1}^\rt P(\vy_{\tau,L}) \vpi_{\tau-L-1}}
- \dfrac{\partial \ln p(\theta)}{\partial \theta_i}.
\end{align}
%\begin{align}
%\label{eq:gradU_single}
%    \widetilde{\nabla U}(\theta)_i =
%	- {\nabla[\ln p(\vy | \theta)]^{\vy_{\tau,L}}_i}
%    - \dfrac{\partial \ln p(\theta)}{\partial \theta_i}.
%\end{align}
To control the variance of this estimator, we sample a collection
of subsequences---referred to as a \emph{minibatch}---$\widetilde{S} = \{\vy_{\tau,L}\}$ where $|\widetilde{S}|$ denotes the
number of subchains in the minibatch. The $\tau$s are drawn randomly from $\{L+1,\ldots,T-L-1\}$; details of the full sampling scheme for $\widetilde{S}$ is provided in the Supplement. We then use the following estimator of the full gradient:
\begin{align}
	&\widetilde{\nabla U}(\theta)_i = \label{eq:gradU_sub} \\\nonumber
    &-\dfrac{1}{p(\widetilde{\mathcal{S}})}\sum_{\vy_{\tau,L} \in \widetilde{S}}
    \dfrac{\vq^\rt_{\tau+L+1} \dfrac{\partial P(\vy_{\tau,L})}{\partial
    \theta_i} \vpi_{\tau-L-1}}{\vq^\rt_{\tau+L+1} P(\vy_{\tau,L}) \vpi_{\tau-L-1}} - \dfrac{\partial \ln p(\theta)}{\partial \theta_i}.
\end{align}
%\begin{align}
%\label{eq:gradU_sub}
%	\widetilde{\nabla U}(\theta)_i =
%    -\dfrac{1}{p(\widetilde{\mathcal{S}})}\sum_{\vy_{\tau,L} \in \tilde{S}}
%    {\nabla[\ln p(\vy | \theta)]^{\vy_{\tau,L}}_i} - \dfrac{\partial \ln p(\theta)}{\partial \theta_i}.
%\end{align}
If we sample $\tilde{S}$ from the set of all possible length-$L$ subsequences with probability ${p(\widetilde{\mathcal{S}})} = |\widetilde{S}|L T^{-1}$, then $\mathbb{E}\left[\widetilde{\nabla U}(\theta)\right] = \nabla{U}(\theta)$~\citep{Gopalan}.

Unfortunately, even this stochastic estimate is prohibitively expensive to compute since the $\vq$ and $\vpi$ terms require touching nearly all of the observations. We instead consider approximating these quantities. % using forward and backward messages.

%%% OLD VERSION
%In order to use subsequences to reduce the amount of computation when computing $\nabla U(\theta)$, we inevitably introduce error at the boundaries of the subsequences where dependencies between observations are broken, as shown in Fig.~\ref{fig:cartoon}. In terms of Eq.~\eqref{eq:gradU_full}, we do not have the exact values of $\vq_{\tau+L+1}$ and $\vpi_{\tau-L-1}$, which would be prohibitively expensive to compute, so we instead approximate these terms.

%\paragraph{Gradient Computation with Subsequences}
\paragraph{Approximating messages via buffering}
Inspired by recent work on stochastic variational inference for
HMMs~\citep{Foti:Xu:Laird:Fox:2014},  we introduce a \textit{buffer} of length
$B$ on either end of each subsequence $(\vy_{LB},\vy_{\tau,L},\vy_{RB})$ where
$\vy_{LB} = (y_{\tau-L-B},\ldots,y_{\tau-L-1})$ and $\vy_{RB} =
(y_{\tau+L+1},\ldots,y_{\tau+L+B})$. See Fig.~\ref{fig:cartoon}.
For an irreducible and aperiodic Markov chain, a sufficiently long buffer will render the observations within $\vy_{\tau,L}$ and those outside the buffers approximately independent. This lets us approximate the boundary terms in Eq.~\eqref{eq:gradU_single} as
\begin{equation}
\label{eq:hmm_msg_approx}
\begin{aligned}
	\vpi_{\tau-L-1} &\approx \tilde{\vpi}_{\tau-L-1} = \underbrace{P(y_{\tau-L-1})A\cdots P(y_{\tau-L-B})A}_{P(\vy_{LB})}\vpi_0 \\
	\vq_{\tau+L+1}^\rt &\approx \tilde{\vq}_{\tau+L+1}^\rt = \mathbf{1}^\rt \underbrace{P(y_{\tau+L+B})A \cdots P(y_{\tau+L+1})}_{P(\vy_{RB})}.
\end{aligned}
\end{equation}
Notice that we plug in $\vpi_0$ and $\vone^\rt$ as the initial conditions for the buffers in Eq.~\eqref{eq:hmm_msg_approx}. Though this introduces errors into the computations of $P(\vy_{LB})$ and $P(\vy_{RB})$, these errors will be nearly forgotten for observations in the subchain of interest $\vy_{\tau,L}$ due to the mixing of the underlying Markov chain.
We rewrite the terms in Eq.~\eqref{eq:gradU_full} as
\begin{align}
	%\dfrac{\partial U(\theta)}{\partial \theta} \approx
    \dfrac{\mathbf{1}^\rt P(\vy_{RB})\dfrac{\partial P(\vy_{\tau,L})}{\partial
    \theta}P(\vy_{LB})\vpi_0}{\mathbf{1}^\rt
    P(\vy_{RB})P(\vy_{\tau,L})P(\vy_{LB})\vpi_0}.
\end{align}

%we sample a collection
%of subsequences, $\tilde{S} = \{\vy_{\tau,L}\}$ where $|\widetilde{S}|$ denotes the
%number of subchains. The $\tau$s are drawn randomly from $\{L+1,\ldots,T-L-1\}$. We then use the following estimator of the full gradient
%%\begin{align}
%%\label{eq:gradU_sub}
%%	&\dfrac{\partial \tilde{U}(\theta)}{\partial \theta_i} = \\
%%	&-\dfrac{T}{L|\tilde{S}|}\sum_{\vy_\tau\in \tilde{S}} \dfrac{\mathbf{1}^\rt P(y_{RB}) \dfrac{\partial P(y_\tau)}{\partial \theta_i} P(y_{LB}) \vpi_0}{\mathbf{1}^\rt P(y_{RB}) P(y_\tau) P(y_{LB}) \vpi_0} - \dfrac{\partial \ln p(\theta)}{\partial \theta_i}
%%\end{align}
%\begin{align}
%	&\dfrac{\partial \tilde{U}(\theta)}{\partial \theta_i} = \label{eq:gradU_sub} \\\nonumber
%    &-\dfrac{1}{p(\widetilde{\mathcal{S}})}\sum_{\vy_{\tau,L} \in \tilde{S}}
%    \dfrac{\mathbf{1}^\rt P(\vy_{RB}) \dfrac{\partial P(\vy_{\tau,L})}{\partial
%    \theta_i} P(\vy_{LB}) \vpi_0}{\mathbf{1}^\rt P(\vy_{RB}) P(\vy_{\tau,L}) P(\vy_{LB}) \vpi_0} - \dfrac{\partial \ln p(\theta)}{\partial \theta_i},
%\end{align}
%where if we sample $\vy_{\tau,L}$ uniformly from $\mathcal{S}$,
%${p(\widetilde{\mathcal{S}})} = \dfrac{|\tilde{S}|L}{T}$, so that $\mathbb{E}\left[\nabla\tilde{U}(\theta)\right] = \nabla{U}(\theta)$~\citep{Gopalan}.
We note that Eq.~\eqref{eq:gradU_sub} is computed in time $O(|\widetilde{S}|LK^2)$ using buffers.
When $|\widetilde{S}|L \ll T$ this results in significant computational speedups
over batch inference algorithms.

A critical question that needs to be answered is \emph{how long should the buffers be?} Though previous theory exists to quantify the buffer length, the resulting lengths are often longer than the entire sequence~\cite{LeGland:1997}.
A heuristic solution was suggested~\citep{Foti:Xu:Laird:Fox:2014}, but theoretical justification was lacking. We propose estimating the buffer length using the \textit{Lyapunov exponent} of the \textit{random dynamical system} specified by $A$ and $P(\vy_{\tau,L})$.
The Lyapunov exponent $\mathfrak{L}$ measures the evolution of the distance
between vectors after applying the operator $(P(\vy_{\tau,L})A)[\cdot]$~\cite{RDS}.
By generalizing the Perron--Frobenius theorem, all of the eigenvalues of
the operator $(P(\vy_{\tau,L})A)[\cdot]$ are less than $0$, which implies that
$\mathfrak{L} \leq 0$~\cite{MarkovChains}. The greater the absolute value of $\mathfrak{L}$, the
faster the errors at the boundaries of the buffers decay, and the shorter the buffers need to be.
Given an estimate of $\mathfrak{L}$, we set the buffer length as $B = \left \lceil
{\mathfrak{L}}^{-1} \ln\left({\delta}/{\delta_0}\right) \right \rceil$ where
$\delta \leq \delta_0$ are error tolerances.
%In particular, the question of buffer length is equivalent to: for two random vectors $\vpi$ and $\vpi^*$, what's the expected length of $LB$ such that after the operation of $P(\vy_{LB})$, $\vpi$ and $\vpi^*$ will synchronize?
%This is a question of random dynamical systems and can be answered through defining the \textit{Lyapunov exponent} as:
%\begin{align}
%\mathfrak{L} = \int_{\Omega\times\mathbb{R}^{K-1}}\ln|| \nabla_{\vr} F_{\vy}(\vr) || \rd \mu_{\vy} \rd \mu_{\vr},
%\end{align}
%where $\vy\in\Omega$, $\vr$ is $\vpi$ transformed through stereographic projection into $K-1$ dimensions.
%The equivalent dynamics $F_{\vy}$ over $\vr$ is the operator $P(\vy)A$ on the projected space.
%Measure $\mu_{\vy}$ corresponds to the distribution of the data $\vy$, and $\mu_{\vr}$ is the invariant measure of $\vr$ under the dynamics of $P(\vy)A$, which will be estimated through sampling.
The method of calculating $\mathfrak{L}$ is described in the Supplement.
Forthcoming work in the applied probability literature formalizes the validity of this approach \cite{FelixYe}.

\paragraph{Approximately independent subsequences}
We estimate $\widetilde{\nabla U}(\theta)$ with minibatches composed of subsequences.
When the subsequences in a minibatch overlap or are very close to one another, the statistical efficiency of $\widetilde{\nabla U}(\theta)$ is diminished,
% used to estimate Eq.~\eqref{eq:gradU_sub} are sampled naively, they will often overlap which diminishes the statistical efficiency of
%the estimator,
requiring more subsequences to obtain accurate estimates. If we
assume that the Markov chain of the latent state sequence is in equilibrium ---
a realistic assumption if $T$ is huge --- then we can leverage the memory decay of the Markov chain to encourage independent subsequences for use in the gradient estimator.

The mixing time of a Markov chain, denoted $\nu$, is the number of steps needed until the chain is ``close" to its stationary distribution~\citep{MarkovChains}. This implies that for $|t - \tau| > \nu$, the corresponding $x_t$ and $x_{\tau}$ are approximately independent. Consequently, if we choose the buffer length $B>0$ s.t. $L+B \geq \nu$, then $t < \tau - L -B$ or $t > \tau + L + B$ implies that $y_t$ is approximately independent of $\vy_{LB}$, $\vy_{\tau,L}$, and $\vy_{RB}$. Therefore, we can increase the statistical efficiency of $\widetilde{\nabla U}(\theta)$ by sequentially sampling the $\vy_{\tau,L}$s such that they are at least $2(L+B)+\nu$ time indices apart (see the Supplement for details).
%
%
%\nf{This changes the scaling factor necessary to make the gradient estimate unbiased. We should specify what this constant is and reference Gopalan's SVI for relational models paper.}
We estimate the mixing time $\nu = (1 - \hat{\lambda}_2)^{-1}$ where $\hat{\lambda}_2$ is the second largest eigenvalue of the current transition parameter iterate, $A^{(t)}$.

When sampling subsequences adhering to the mixing-time-dependent gap, each term in Eq.~\eqref{eq:gradU_sub} is rendered approximately independent. Following standard practice for SG-MCMC algoriths, we appeal to the central limit theorem obtaining the following expression for the asymptotic distribution of $\widetilde{\nabla U}(\theta)$ as:
\begin{equation}
	\widetilde{\nabla U}(\theta) \approx \nabla U(\theta) + \mathcal{N}(0, V_i(\theta)),
	\label{eq:SGMCMCHMM_CLT}
\end{equation}
where $V_i(\theta)$ is the stochastic gradient noise variance. This will prove crucial for our analysis in Sec.~\ref{sec:analysis}.
%\nf{Transition sentence connecting with correctness of SG-MCMC from background, but that novel error sources arose.}

%% MOVED SOURCES OF ERROR FROM HERE!!!

%Both of these challenges are inherent to standard SG-MCMC methods for i.i.d.\ data.
%When the step size $\epsilon_t$ decreases to zero, the asymptotic bias from the error sources are asymptotically $0$.
%This bias-variance trade-off was recently studied in~\cite{Vollmer}.
%Higher order numerical schemes \cite{DingHighOrder,Shang_new} and a moving window estimation of $\widehat{B}$ can further reduce this bias \cite{Shang}.
%Similar to the suggestions therein, we recommend that when the number of minibatches used in each gradient calculation is less than the order of $100$, simply take a small (or decreasing) stepsize instead of estimating $\widehat{B}$ from the subsampled data, as a reliable estimate may not be available from the data;
%when the number of minibatches exceeds the order of $100$, the central limit theorem takes effect and one can estimate the stochastic gradient noise $\widehat{B}$ and subtract it from the injected noise.
%A moving window estimate can further reduce this bias \cite{Shang}.
%
%Having characterized the error introduced by our approximations, we can use the results from~\citet{completesample,complete_new} to show that the proposed SG-MCMC for HMMs asymptotically has the right stationary distribution.
%\nf{expand a bit connecting to results in background}

\subsection{Incorporating Geometric Information}

%Though the previously derived algorithm theoretically attains the correct stationary distribution, in practice we fine that it suffers from some numerical problems.
Eq.~\eqref{eq:update} serves as a general purpose algorithm that theoretically attains the correct stationary distribution for any $\mD$ and $\mQ$ matrices when the step size $\epsilon_{t}$ approaches zero.
But in practice, we need to take into account numerical stability during numerical integrals.
For example, when we are sampling from the probability simplex, previous work has shown that taking the curvature of the parameter space into account is important~\citep{SGLD,completesample}.
Since our transition parameters live on the simplex, we likewise incorporate the geometry of the parameter space by constructing a \textit{stochastic-gradient Riemannian MCMC} (SG-RMCMC) algorithm.
%We sample transition parameters and emission parameters iteratively, using mean estimates obtained from previous sampling procedure as fixed parameters in the next one.

\paragraph{SG-RLD for transition parameters}
In order to sample the transition matrix $A$ we note that the columns of $A$ are constrained to lie on the probability simplex.
To address these constraints, we use the expanded mean parametrization:
$A = \dfrac{|\hat{A}_{i,j}|}{\sum_{i}|\hat{A}_{i,j}|}$, similar to
what \citet{SGRLD} used for topic modeling.
Evaluating $\widetilde{\nabla U}(\theta)$ in Eq.~\eqref{eq:gradU_full} for $\theta = \hat{A}_{i,j}$, using Eq.~\eqref{eq:hmm_subchain_lik} yields:
%The gradient for the entries of the transition matrix, $A_{i,j}$ is given by
%When $\theta_i \in \{A_{i,j}\}$,
\begin{align}
&\widetilde{\nabla U}_{\hat{A}_{i,j}}(\hat{A}_{i,j}) \label{eq:gradU_A}  \\ \nonumber
=
    &- \dfrac{1}{p(\widetilde{\mathcal{S}})} \sum_{\vy_{\tau,L}\in \widetilde{\mathcal{S}}}
\sum_{t=\tau-L}^{\tau+L}
%\dfrac{ \mathbf{1}^\rt M_\tau^{LB}
\dfrac{ \left(\widetilde{\vq}_{\tau+L+1}\right)_i P_{i,i}(y_t) \left(\widetilde{\vpi}_{\tau-L-1}\right)_j}
{\widetilde{\vq}_{\tau+L}^\rt P(y_t) \hat{A} \widetilde{\vpi}_{\tau-L}}.
%\nonumber\\
%&= - \dfrac{T}{l\cdot |\widetilde{\mathcal{S}}|} \sum_{\tau\in \widetilde{\mathcal{S}}}
%\sum_{t=\tau-L}^{\tau+L}
%\dfrac{\widetilde{p}_\tau (x_t = i, x_{t-1} = j | \vy_\tau)}{A_{i,j}},
\end{align}
Here, $\tilde{\vpi}_{\tau-L-1}$ and $\tilde{\vq}_{\tau+L+1}$ are computed on the left and right buffers, respectively, according to Eq.~\eqref{eq:hmm_msg_approx}. The terms inside the sum in Eq.~\eqref{eq:gradU_A} are analogous to the pairwise marginals of the latent state in traditional HMM inference algorithms.
A detailed derivation of this gradient can be found in the Supplement.
%where $\widetilde{p}_\tau$ is calculated on $\vy_{LB}$, $\vy_\tau$, and $\vy_{LB}$.
%\nf{Probably need a bit more detail on how $\tilde{p}_\tau$ is computed since we integrate out the $x_t$s.}

By leveraging the flexible SG-MCMC update rule of Eq.~\eqref{eq:true-update}, we remove the dependency on $\hat{A}_{i,j}$ from the denominator of Eq.~\eqref{eq:gradU_A} by selecting $\mD = \hat{A}$ and $\mQ = \mathbf{0}$.
%This yields a two-step update:
%Since the parameter of interest $A_{i,j}$ shows up in the denominator we utilize the flexible SG-MCMC update rule in Eq.~\eqref{eq:true-update} and set $\mD = A$ and $\mQ = \mathbf{0}$, which results in the following update rule that can be used in Alg.~\ref{alg:SG_MCMC}:
%We make use of the update rule in Eq.~\eqref{eq:true-update} and Alg.~\ref{alg:SG_MCMC} to sample parameters $A$.
%We take $\mD = A$, and $\mQ = {\bf 0}$ in the stochastic dynamics, resulting in the update rule:
%We further use expanded mean parametrization of $A$:
This yields the following update:
\begin{align}
\hat{A}_{i,j}^{(t+1)} \leftarrow& \hat{A}^{(t)}_{i,j} -  \eps_t \left[\hat{A}^{(t)}_{i,j} \widetilde{\nabla U}_{\hat{A}_{i,j}}(\hat{A}^{(t)}_{i,j}, \phi) + I \right] \nonumber\\
& + \mathcal{N}\left (0, \eps_t \left (2 \hat{A}^{(t)}_{i,j} - \eps_t \widehat{\mB}^{(t)}_{i,j} \right ) \right)
\label{eq:update_A}
\end{align}
%\begin{equation}
%\begin{aligned}
%\hat{A}_{i,j} \leftarrow& A^{(t)}_{i,j} -  \eps_t \left[A^{(t)}_{i,j} \nabla \widetilde{U}(A^{(t)}_{i,j}, \phi) + I \right] \nonumber\\
%& + \mathcal{N}\left (0, \eps_t \left (2 A^{(t)}_{i,j} - \eps_t \widehat{\mB}^{(t)}_{i,j} \right ) \right) \nonumber \\
%\label{eq:update_A}
%A^{(t+1)} &= \dfrac{|\hat{A}_{i,j}|}{\sum_{i}|\hat{A}_{i,j}|},
%\end{aligned}
%\end{equation}
where $\phi$ denotes all other model parameters.
%The first step of the update is the proposed SGR-MCMC update. However, the noise introduced by the Langevin dynamics can result in values of $A_{i,j}$ that are less than zero or greater than one. As such, we actually perform the SGR-MCMC update on the \textit{expanded mean} parameterization of $A$~\cite{} (REF NEEDED). The second step of the update projects the new iterate back onto the space of transition probabilities.
We note that this pre-conditioned gradient takes advantage of the local geometry of the parameter space by pre-multiplying by a metric tensor that arises from Eq.~\eqref{eq:true-update}.

\paragraph{SG-RLD for emission parameters}
Similarly to the transition parameters, we sample the emission parameters $\{\phi_k : k=1,\ldots,K\}$, by evaluating $\widetilde{\nabla U}(\theta)$ in Eq.~\eqref{eq:gradU_full} for $\theta = \phi_k$ Using Eq.~\eqref{eq:hmm_subchain_lik}. This results in the gradient:
%we obtain the gradient
%When $\theta_i \in \{\phi_k\}$,
\begin{align}
\widetilde{\nabla U}(\phi_k)
    =& - \dfrac{1}{p(\widetilde{\mathcal{S}})} \sum_{\vy_{\tau,L}\in \widetilde{\mathcal{S}}}
\sum_{t=\tau-L}^{\tau+L}
\\ \label{eq:gradU_Emission}
&\dfrac{ \left(\widetilde{\vq}_{\tau+L+1}\right)_k P_{k,k}(y_t) \left(\widetilde{\vpi}_{\tau-L-1}\right)_k}
{\widetilde{\vq}_{\tau+L+1}^\rt P(y_t) A \widetilde{\vpi}_{\tau-L-1}}
\cdot \dfrac{\partial \ln P_{k,k}(y_t)}{\partial \phi_k} \nonumber.
%=& - \dfrac{T}{l\cdot |\widetilde{\mathcal{S}}|} \sum_{\tau\in \widetilde{\mathcal{S}}}
%\sum_{t=\tau-L}^{\tau+L}
%\widetilde{p}_\tau (x_t = k | \vy)
%\cdot \dfrac{\partial \ln P_{k,k}(y_t)}{\partial \phi_k},
\end{align}
Again, $\tilde{\vpi}_{\tau-L-1}$ and $\tilde{\vq}_{\tau+L+1}$ are computed on the left and right buffers, respectively, according to Eq.~\eqref{eq:hmm_msg_approx}.
%where $\widetilde{p}_\tau$ is again calculated from $\vy_{LB}$, $\vy_\tau$, and $\vy_{LB}$.
Similarly to the transition parameters, we account for the geometry of the parameter space by specifying an appropriate $\mD$ and $\mQ$ in Eq.~\eqref{eq:true-update} which in general depends on the form of $p(y_t|\phi)$. For exponential family emission distributions we recommend taking $\mD$ to be the inverse of the \textit{Fisher information matrix}~\cite{Amari:1998}.
%We make use of the update rule in Eq.~\eqref{eq:true-update} and Alg.~\ref{alg:SG_MCMC} to sample parameters $\phi$.
%We take $\mD = A$, and $\mQ = {\bf 0}$ in the stochastic dynamics, resulting in the update rule:
%\begin{align}
%A_{t+1} \leftarrow& A_{t} -  \eps_t \left[A_t \nabla \widetilde{U}(A_t, \phi) + I \right] \nonumber\\
%& + \mathcal{N}(0, \eps_t (2 A_t - \eps_t \widehat{\mB}_t)),  \label{eq:update_A}
%\end{align}
%where $A_t \nabla \widetilde{U}(A_t, \phi) = \widetilde{p}_\tau (x_t = i, x_{t-1} = j | \vy)$ is obtained from the forward backward algorithm on the subchain and the buffer chains.

As a concrete example, we consider a Gaussian emission distribution. Define, $z_t = (\mu_k - y_t)$, then we have:
\begin{align}
\nabla_{\mu_k} \ln P_{k,k}(y_t) &= \Sigma_k^{-1} z_t \\%(\mu_k - y_t) \\
\nabla_{\Sigma_k} \ln P_{k,k}(y_t) &= \dfrac{1}{2} %\Sigma_k^{-1} \left(\Sigma_k - (\mu_k - y_t)(\mu_k - y_t)^\rt \right) \Sigma_k^{-1}.
    \Sigma_k^{-1} \left(\Sigma_k - z_t z_t^\rt \right) \Sigma_k^{-1}.
\end{align}
We plug these values into the SG-MCMC update of Eq.~\eqref{eq:true-update} using $\mD = \Sigma$ to account for the geometry of the parameter space and $\mQ = \mathbf{0}$. This leads to the update equations:
%We thus use the preconditioned gradients $\Sigma_k \nabla_{\mu_k} \widetilde{U}(A, \phi_k(t))$ to update each $\mu_k$ and $\Sigma_k \nabla_{\Sigma_k} \widetilde{U}(A, \phi_k(t)) \Sigma_k$ to update each $\Sigma_k$ in Eq.~\eqref{eq:true-update} yielding the following updates:
\begin{align}
\mu_k^{(t+1)} \leftarrow& \mu_k^{(t)} -  \eps_t \left[\Sigma_k^{(t)} \widetilde{\nabla_{\mu_k} U}(A, \phi_k^{(t)})\right] \nonumber\\
& + \mathcal{N}(0, \eps_t (2 \Sigma_k^{(t)} - \eps_t \widehat{\mB}_t)).  \label{eq:update_mu}
\end{align}
\begin{align}
\Sigma_k^{(t+1)} \leftarrow& \Sigma_k^{(t)} -  \eps_t \left[\Sigma_k^{(t)} \widetilde{\nabla_{\Sigma_k} U}(A, \phi_k^{(t)}) \Sigma_k^{(t)} + \Sigma_k^{(t)}\right] \nonumber\\
& + \mathcal{N}(0, \eps_t (2 \Sigma_k^{(t)}\otimes\Sigma_k^{(t)} - \eps_t \widehat{\mB}^{(t)})).  \label{eq:update_Sigma}
\end{align}
It is possible when using Eq.~\eqref{eq:update_Sigma} to obtain a $\Sigma^{(t+1)}$ that is not positive definite. In this case we reject the update and set $\Sigma^{(t+1)} = \Sigma^{(t)}$.

\subsection{Analysis of SG-MCMC for HMMs}
\label{sec:analysis}
Our proposed SG-MCMC scheme for HMMs introduces error in three ways.
The first two carry over from the standard i.i.d. setting: (i) discretizing the continuous stochastic dynamics and (ii) estimating the stochastic gradient noise covariance, as discussed in Sec.~\ref{sec:SGMCMCiid}.

The third is via our approximations $\tilde{\vpi}_{\tau-L-1}$ and $\tilde{\vq}_{\tau+L+1}$. This error is already incorporated in Eq.~\eqref{eq:SGMCMCHMM_CLT}, and vanishes with $\epsilon_t$ in Eq.~\eqref{eq:true-update}. Thus, applying the results from~\citet{completesample,complete_new}, we can show that the proposed SG-MCMC for HMMs asymptotically has the right stationary distribution under the same conditions as in the i.i.d. case.  However, in practice we use a fixed $\epsilon_t$, and as we show in Sec.~\ref{exp:buffer}, performing sufficient buffering via our Lyapunov exponent approach is critical.

In summary, our SG-MCMC algorithm enables MCMC-based inference in HMMs for massive sequences of data. In particular, we only require computations on collections of small subsequences and attain the desired stationary distribution by mitigating the errors incurred by these approximations. Finally, we have shown how to incorporate geometric information about the parameter space in order to increase the numerical robustness of the algorithm.

\begin{algorithm}[t!]
\caption{SG-MCMC for HMM}
\begin{algorithmic}
\STATE initialize $A^{(0)}$ and $\phi_k^{(0)}$
 \FOR{$n=0,1,2\cdots,N_{\mathrm{iter}}$}
    \STATE Periodically estimate the buffer length $B$ and the minimum subchain
    gap $\nu$ according to Sec.~\ref{sec:sgmcmc}.
    \STATE Sample subchains $\tilde{S}$ of length $L$ from $p(\widetilde{S})$.
    \FOR{$s=1\cdots N_{\mathrm{steps}}$}
    \STATE
    Update $\hat{A}^{(s)}$ according to Eq.~\eqref{eq:gradU_A} and \eqref{eq:update_A}
    \ENDFOR
    \STATE
    Calculate $\hat{A} = \frac{1}{N_{\mathrm{steps}}} \sum_{s=1}^{N_{\mathrm{steps}}} \hat{A}^{(s)}$.
    \STATE
    Set $A_{i,j} \leftarrow {|\hat{A}_{i,j}|}\Big/{\sum_i |\hat{A}_{i,j}|}$
    \FOR{$s=1\cdots N_{\mathrm{steps}}$}
    \STATE
    update $\phi^{(s)}$ according to Eqs.~\eqref{eq:gradU_Emission}--~\eqref{eq:update_Sigma}
    \ENDFOR
    \STATE
    Set $\phi = \frac{1}{N_{\mathrm{steps}}} \sum_{s=1}^{N_{\mathrm{steps}}} \phi^{(s)}$.
   \ENDFOR
\end{algorithmic}
\label{alg:SG_MCMC}
\end{algorithm}

\section{Experiments}

We evaluate the performance of our proposed SG-RLD algorithm for HMMs on both synthetic and real data.
%To evaluate the algorithm on different models we have used
First, we demonstrate the effectiveness of incorporating the buffers on two synthetic data sets that exhibit very different dynamics. Next we apply our SG-RLD for HMMs to a non-conjugate model of synthetic data. Finally, we apply SG-RLD to a large ion channel recording data set and compare to batch MCMC.

\subsection{Evaluating Buffer Effectiveness}
\label{exp:buffer}
\begin{figure}[t!]
%\twocolumn[
\centering{
\includegraphics[height=0.19\textwidth]{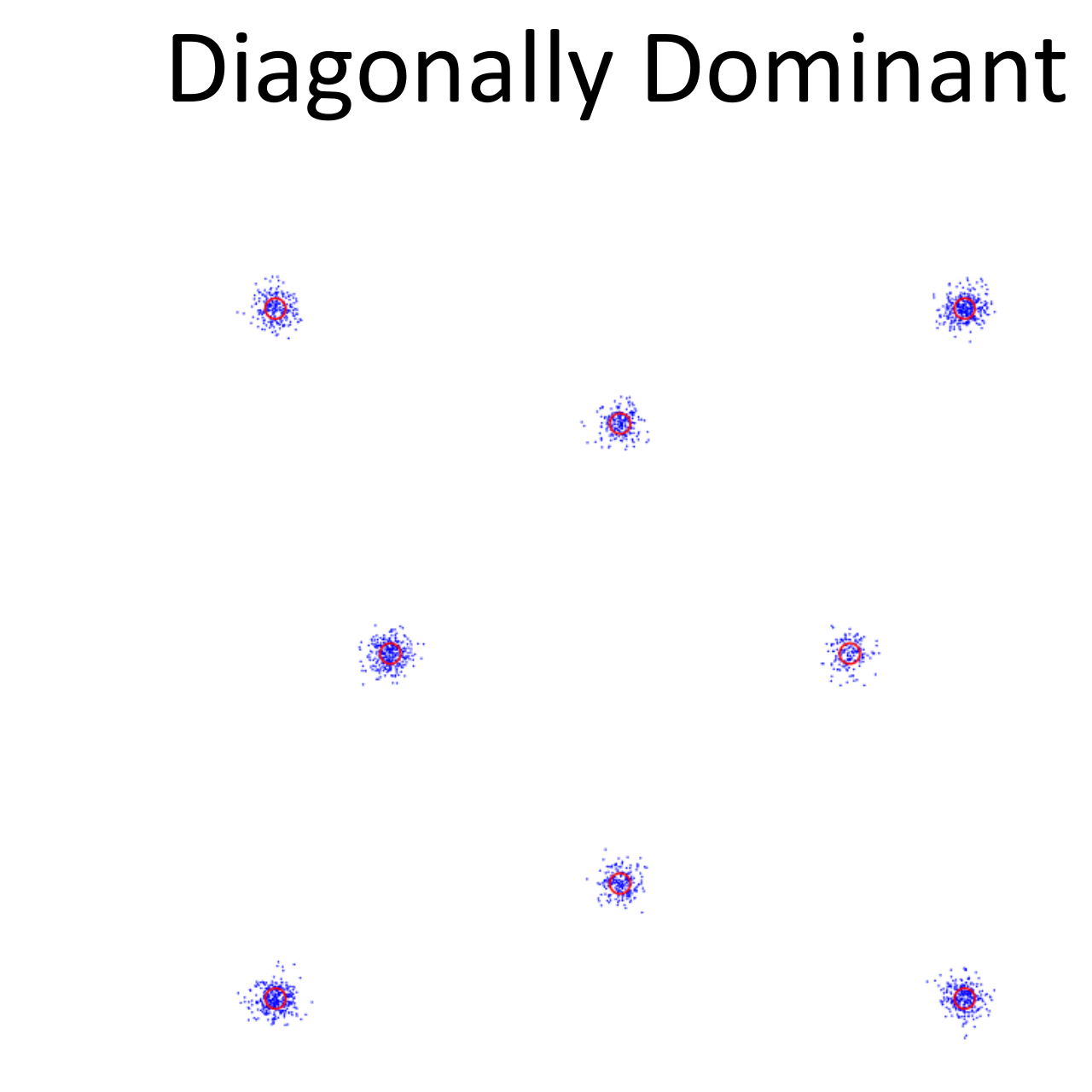}
\includegraphics[height=0.19\textwidth]{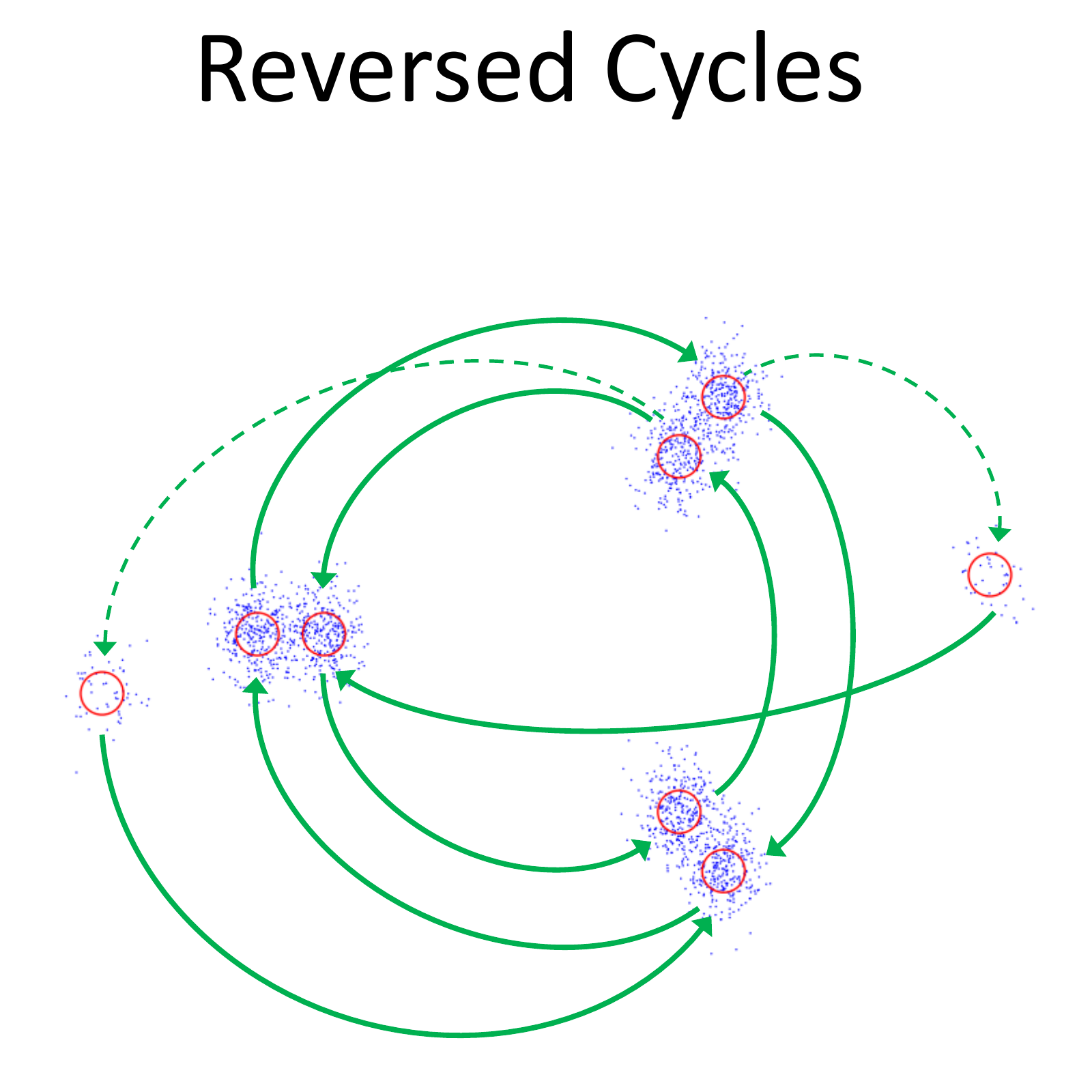}\\
\includegraphics[height=0.185\textwidth]{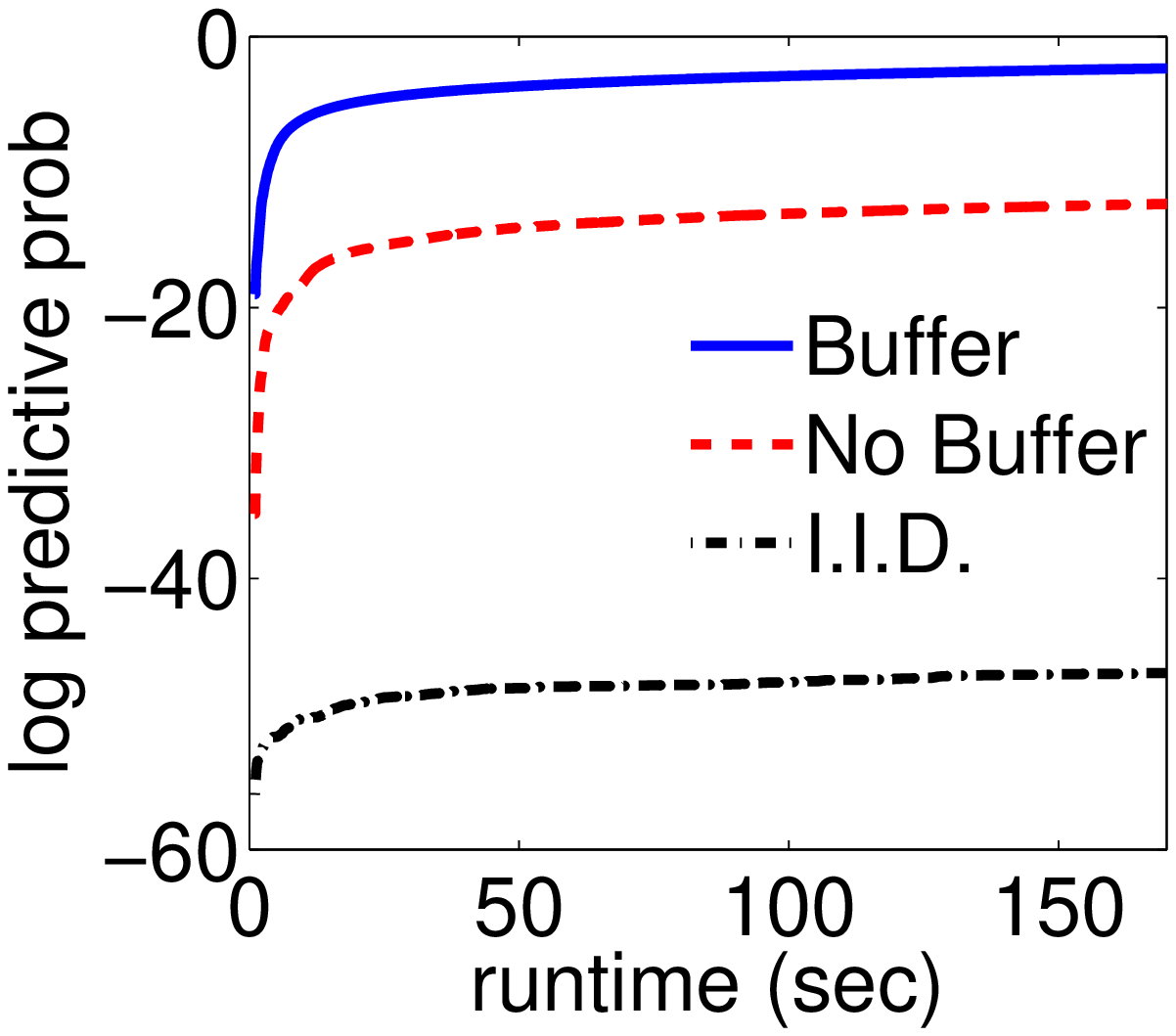}
\includegraphics[height=0.185\textwidth]{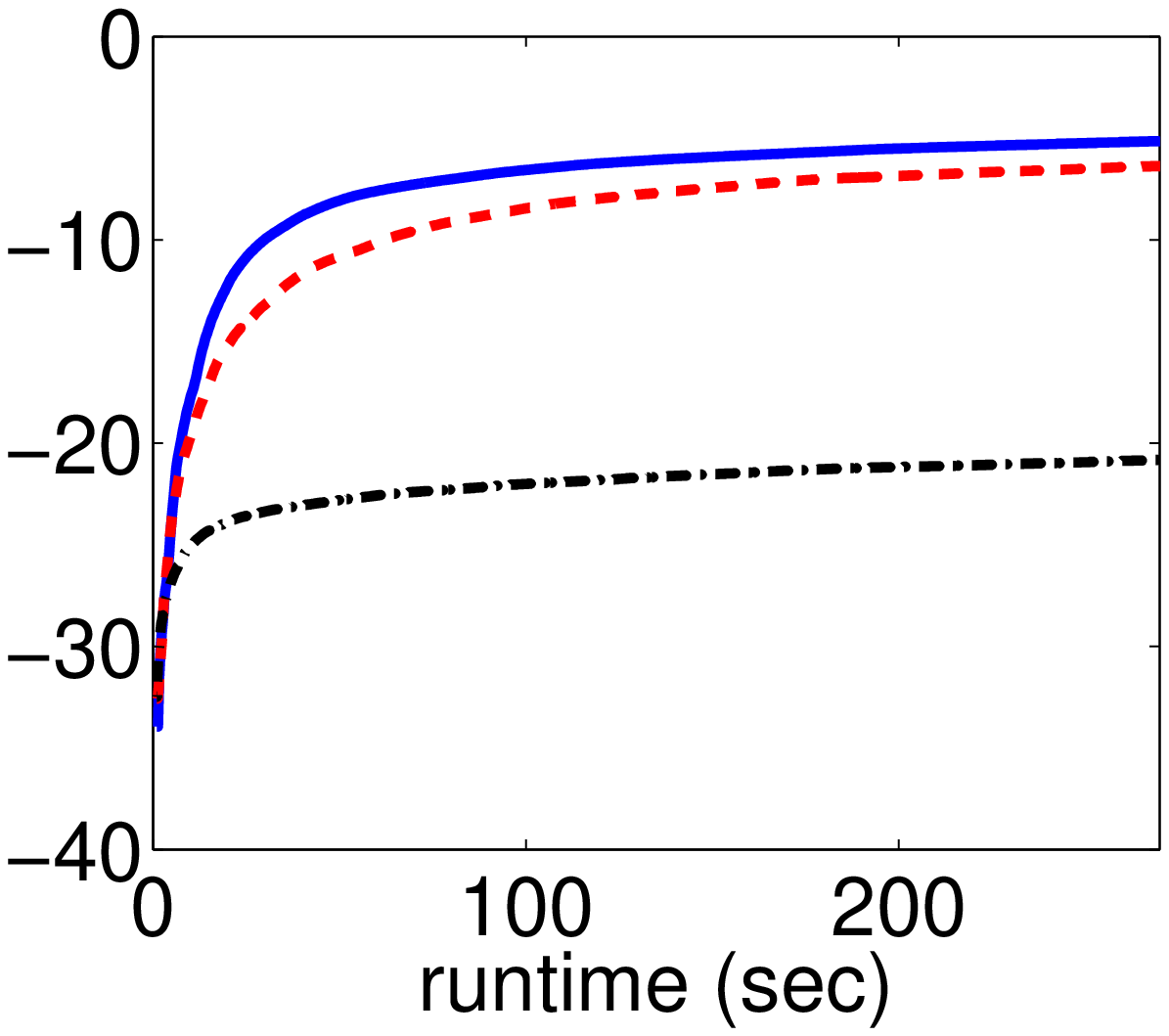}
}
\vspace{-0.1in}
\caption{\small Synthetic experiments for
DD (\emph{left}) and RC (\emph{right}) data.
\emph{Top:} Sample datasets;
Arrows indicate Markov transitions.
\emph{Bottom:} 10-step-ahead log predictive likelihood versus learning time for DD (\emph{left}) and RC (\emph{right}) dynamics.
Comparisons are made for SG-RLD algorithms with adaptive buffer, no buffer, and treating the data as i.i.d..
%All of the experiments use a constant computation budget by varying the number of subchains, $|\tilde{S}|$, with the length of the subchains, $L$.
}\vspace{-0.1in}
%]
\label{fig:sim}
\end{figure}

We first design two synthetic experiments in order to illustrate the effectiveness of our adaptive buffer scheme. We compare SG-RLD with buffering, without buffering, and treating the data as i.i.d.\ as a baseline. We can view the no-buffer algorithm as one that treats the subsequences as short, independent realizations, similarly to~\cite{Johnson:Willsky:2014}. Following~\citet{Foti:Xu:Laird:Fox:2014}, we create two synthetic datasets both with $T = 20$ million observations and $K = 8$ latent states (see Fig.~\ref{fig:sim} (top)).

The first data set, \textit{diagonally dominant} (DD) consists of a Markov chain that heavily self-transitions and has identifiable emissions. %Inferring the emissions is easy in this case, however, learning the transition matrix is difficult due to the rarity of transitions.
The second dataset---\textit{reversed cycles} (RC)---strongly transitions between two cycles over three states, each in opposite directions. Further details on these datasets and how we set $L$ and $|\widetilde{S}|$ are in the Supplement.
%We choose $L=2$ and $|\widetilde{S}|=10$ subsequences in order to incorporate observations from distant parts of the observation sequence.
% We set $L=5$ and $|\widetilde{S}|=4$

The 10-step-ahead predictive log probability is depicted in Fig.~\ref{fig:sim} for the DD and RC datasets. See the Supplement for similar results comparing errors in transition matrix estimation. In both cases, we see that both SG-RLD HMM methods greatly outperform the i.i.d.\ algorithm. The reason i.i.d.\ SG-RLD performs so badly on the DD data stems from all states being equally probable so that ignoring the dynamics forces the model to have little apriori confidence in the next observations. For the RC dataset, the i.i.d.\ algorithm fails to capture the structured transitions between states, again reducing predictive performance. Importantly, our adaptive buffer scheme attains both better predictive performance and converges to the true transition matrix in less time. In fact, there is a bias in the learned transition matrix for the non-buffered algorithm due to inaccurate subchain approximation of $\vq, \vpi$.
This experiment demonstrates that accounting for dynamics yields massive gains in predictive performance and that using our adaptive buffer scheme provides further gains on top of that.
%\begin{tabular}{ccc}
%        \hspace{-0.3in}
%        \vspace{-0.05in}
%        \includegraphics[width = 1.8in]{figure/ShortTraceMHOnTorusNew.eps} & \hspace{-0.2in}
%        \includegraphics[width = 1.8in]{figure/ShortTraceIrrOnTorusNew.eps} & \hspace{-0.2in}
%        \includegraphics[width = 1.8in]{figure/ShortTraceIrrMALAOnTorus.eps} \\ \hspace{-0.2in}
%        \includegraphics[width = 1.8in]{figure/TraceMHOnTorusNew.eps} & \hspace{-0.2in}
%        \includegraphics[width = 1.8in]{figure/TraceIrrOnTorusNew.eps} & \hspace{-0.2in}
%        \includegraphics[width = 1.8in]{figure/TraceIrrMALAOnTorus.eps} \\
%        \hspace{0.25in}
%        \textbf{MH} &
%        \hspace{0.2in}
%        \textbf{Irreversible Jump} &
%        \hspace{0.3in}
%        \textbf{Irreversible MALA}\\
%    \end{tabular}
%TODO update figure caption
%\begin{figure}[t!]
%%\twocolumn[
%\centering{
%\includegraphics[height=0.185\textwidth]{pics/diag_dom_pred_prob}
%\includegraphics[height=0.185\textwidth]{pics/rev_cyc_pred_prob}
%}
%\vspace{-0.1in}
%\caption{\small Synthetic experiments with hard-to-capture dynamics.
%Log predictive likelihood of $10$ unobserved data points versus learning time for diagonally dominant (\emph{left}) and reversed cycles (\emph{right}) dynamics.
%Comparisons are made for SG-RLD algorithms with estimated buffer, without buffer, and treating data as i.i.d.
%%All of the experiments use a constant computation budget by varying the number of subchains, $|\tilde{S}|$, with the length of the subchains, $L$.
%}\vspace{-0.1in}
%%]
%\label{fig:sim}
%\end{figure}
\begin{figure}[t!]
%\begin{minipage}{0.2\textwidth}
\includegraphics[height=0.14\textheight]{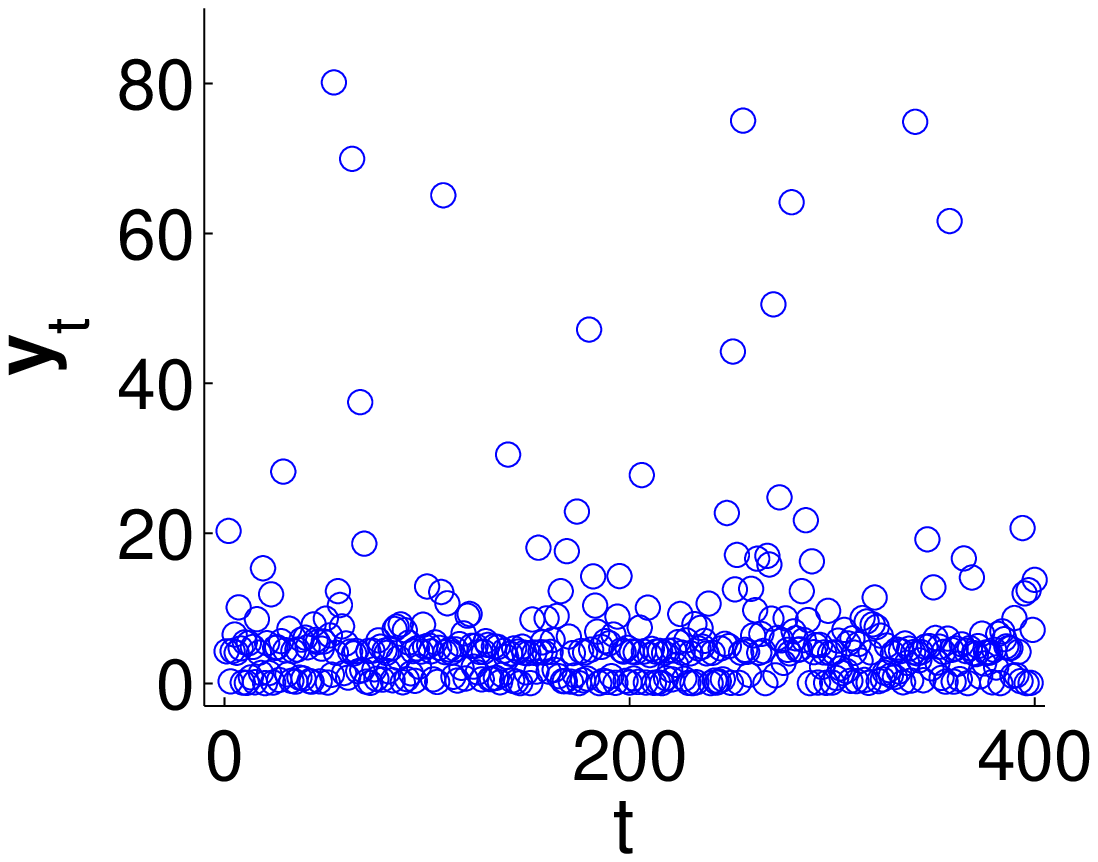}
\includegraphics[height=0.14\textheight]{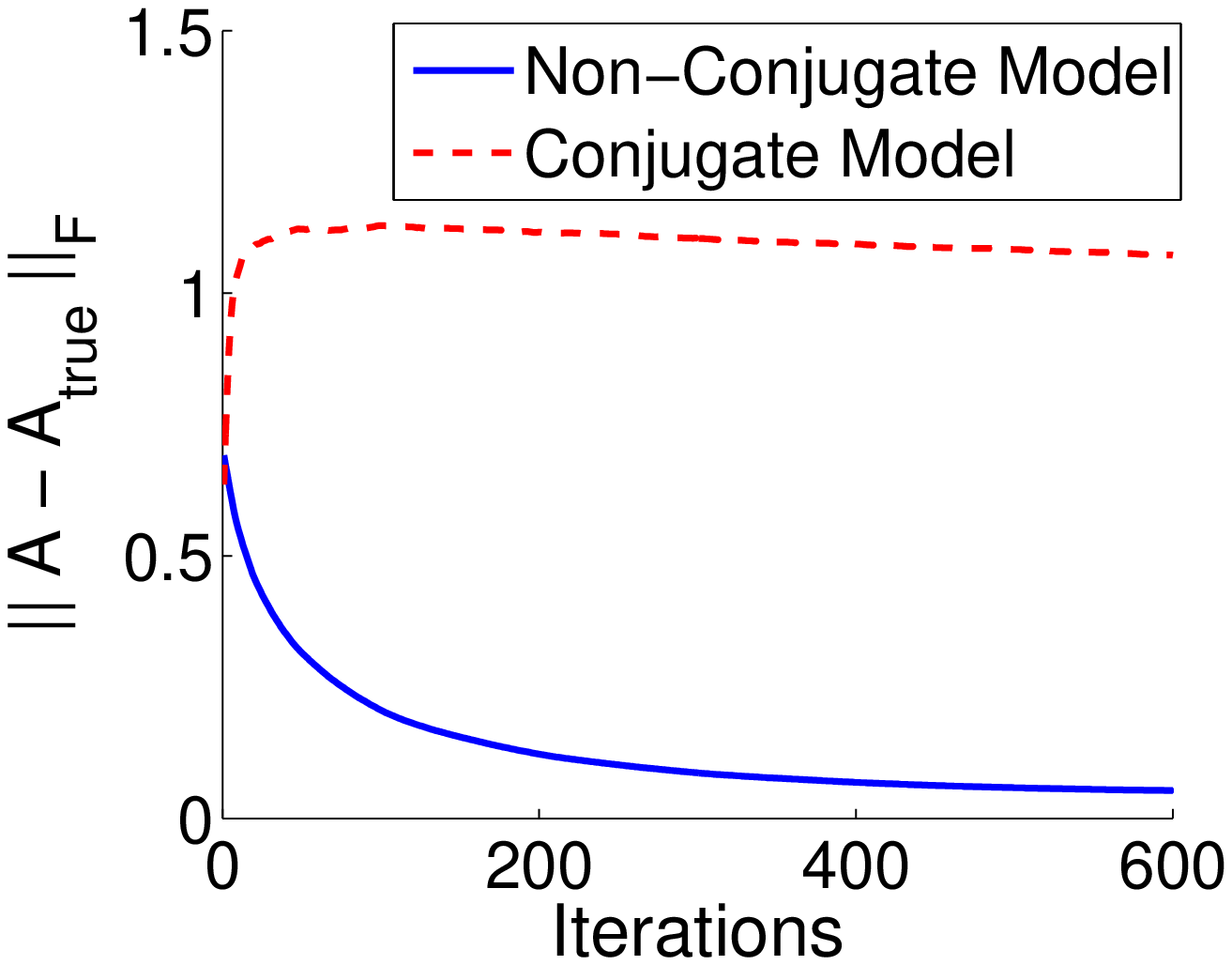}
%\end{minipage}
%\hfill
%\hspace{0.2in}
%\begin{minipage}{0.2\textwidth}
\centering{
\includegraphics[height=0.12\textheight]{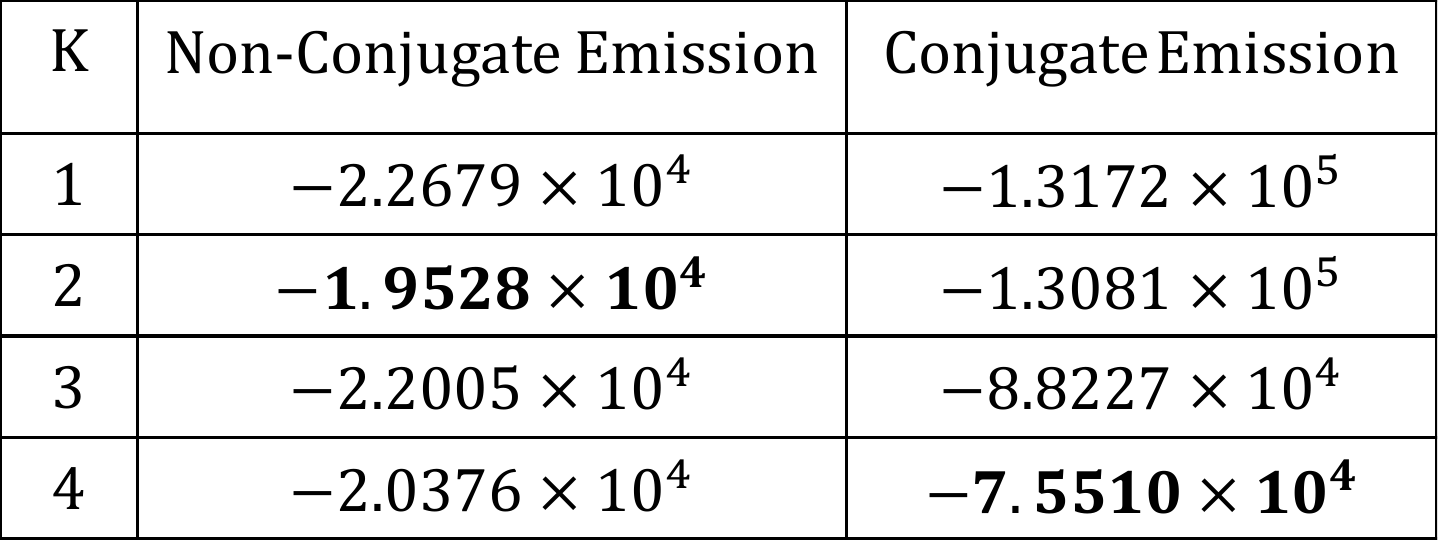}
}
%\end{minipage}
\caption{\small Synthetic experiment with log-normal emission.
We use the non-conjugate emission model on the synthetic data (\emph{Top Left}) with two hidden states and log-normal emissions and compare it against the conjugate model.
We show the difference in convergence speed (\emph{Top Right}) and log held out probability $\ln p(\vy_{\rm test} | \vy_{\rm train})$ on $2,000$ test data (\emph{Bottom}).
}
\vspace{-0.1in}
\label{fig:non-conj}
\end{figure}
\begin{figure*}[t!]
\includegraphics[height=0.14\textwidth]{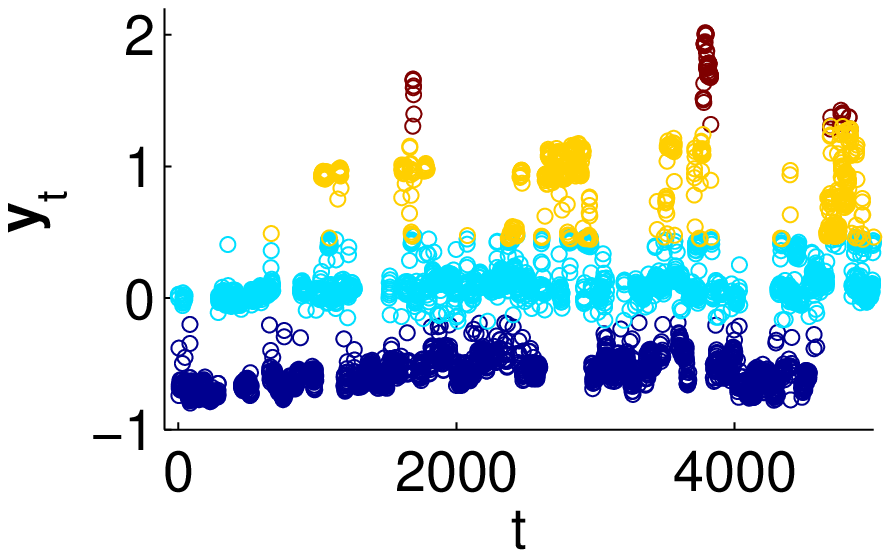}
\includegraphics[height=0.14\textwidth]{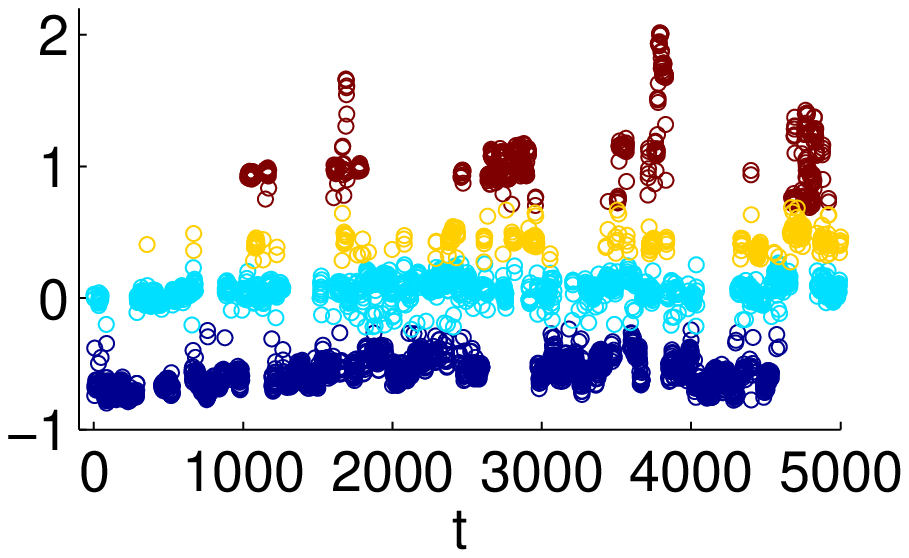}
\includegraphics[height=0.14\textwidth]{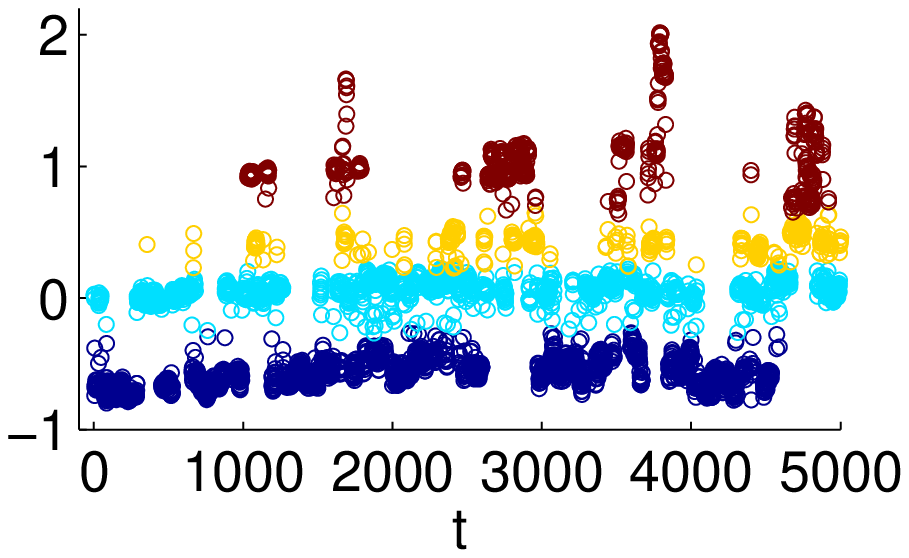}
\includegraphics[height=0.14\textwidth]{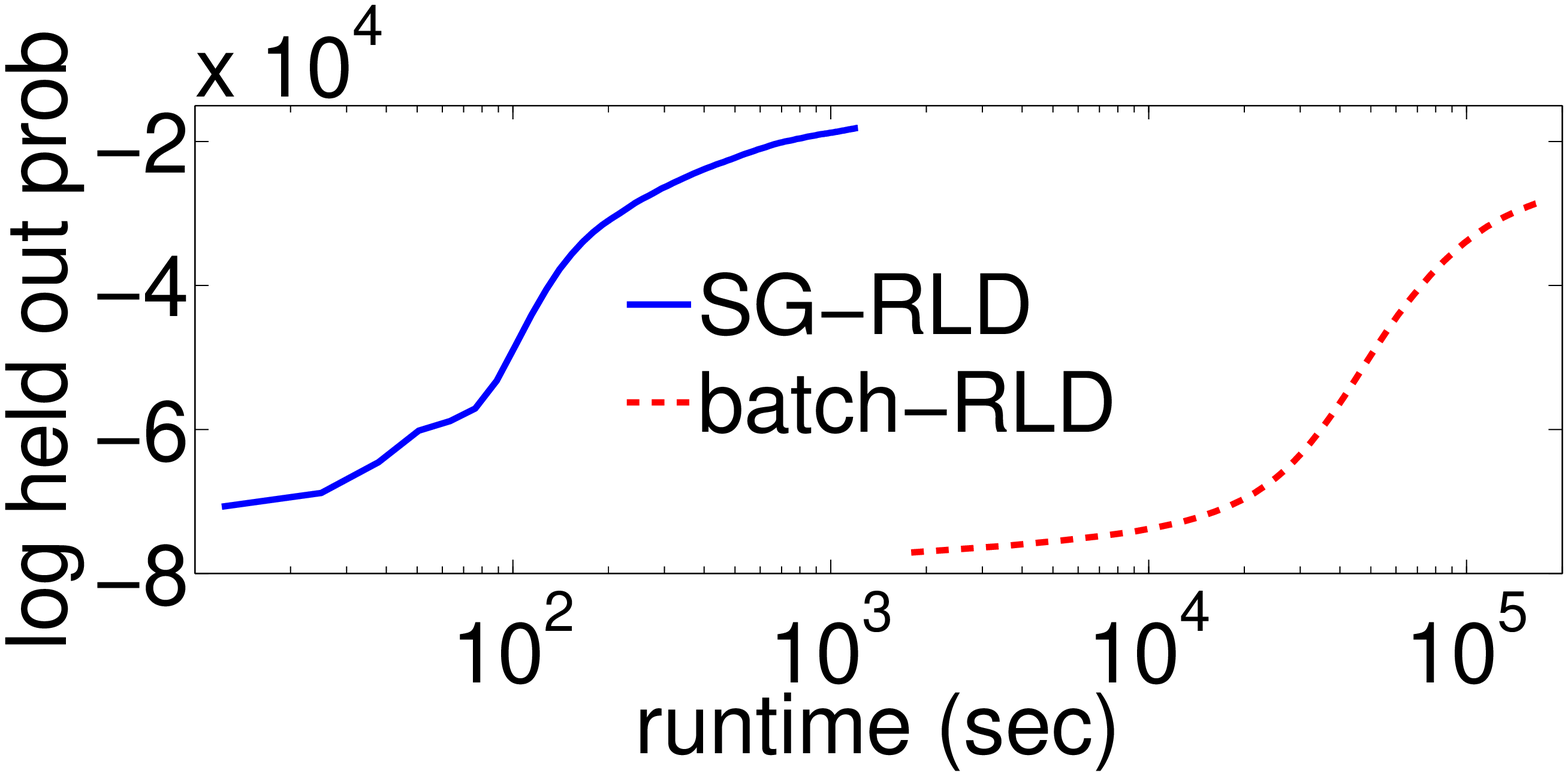}\\
\includegraphics[height=0.14\textwidth]{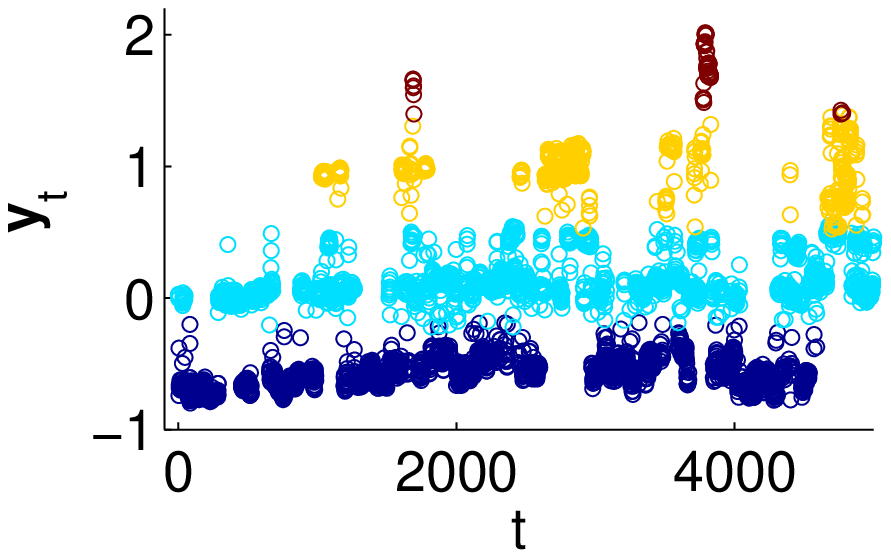}
\includegraphics[height=0.14\textwidth]{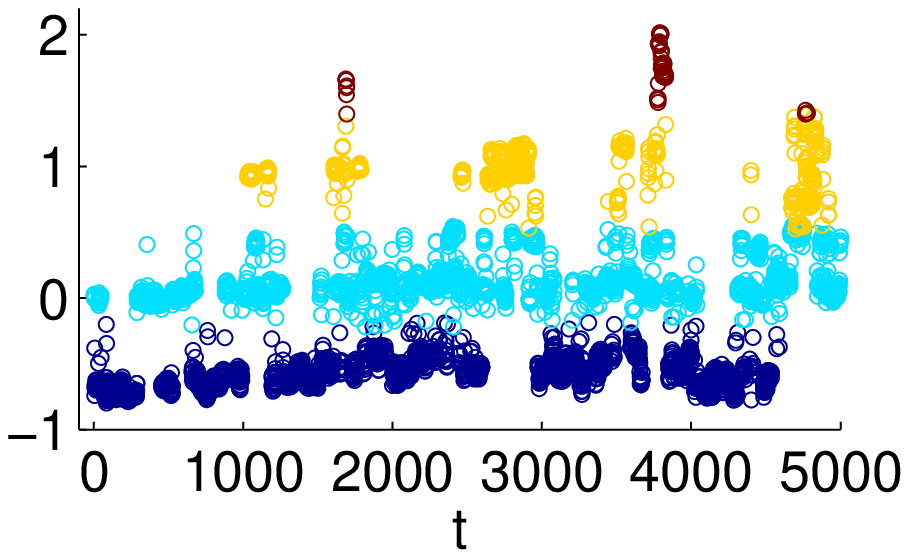}
\includegraphics[height=0.14\textwidth]{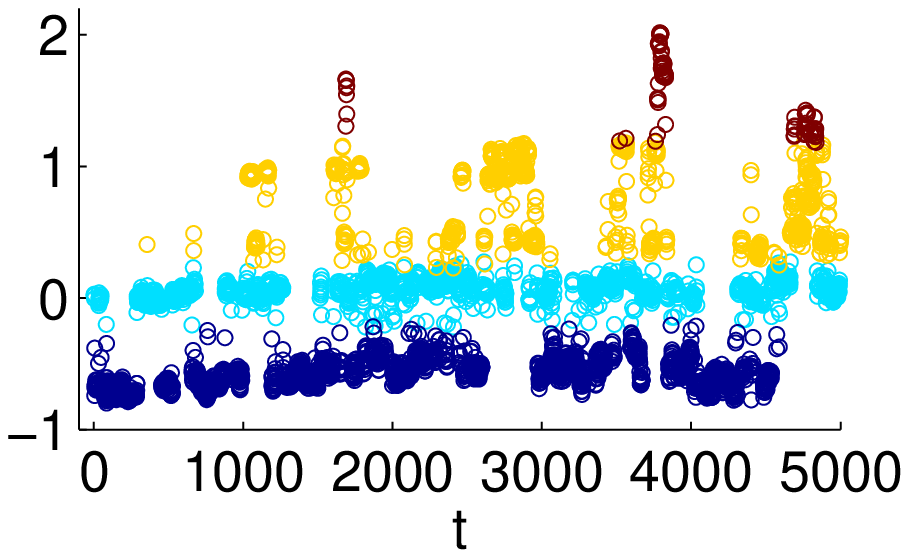}
\includegraphics[height=0.14\textwidth]{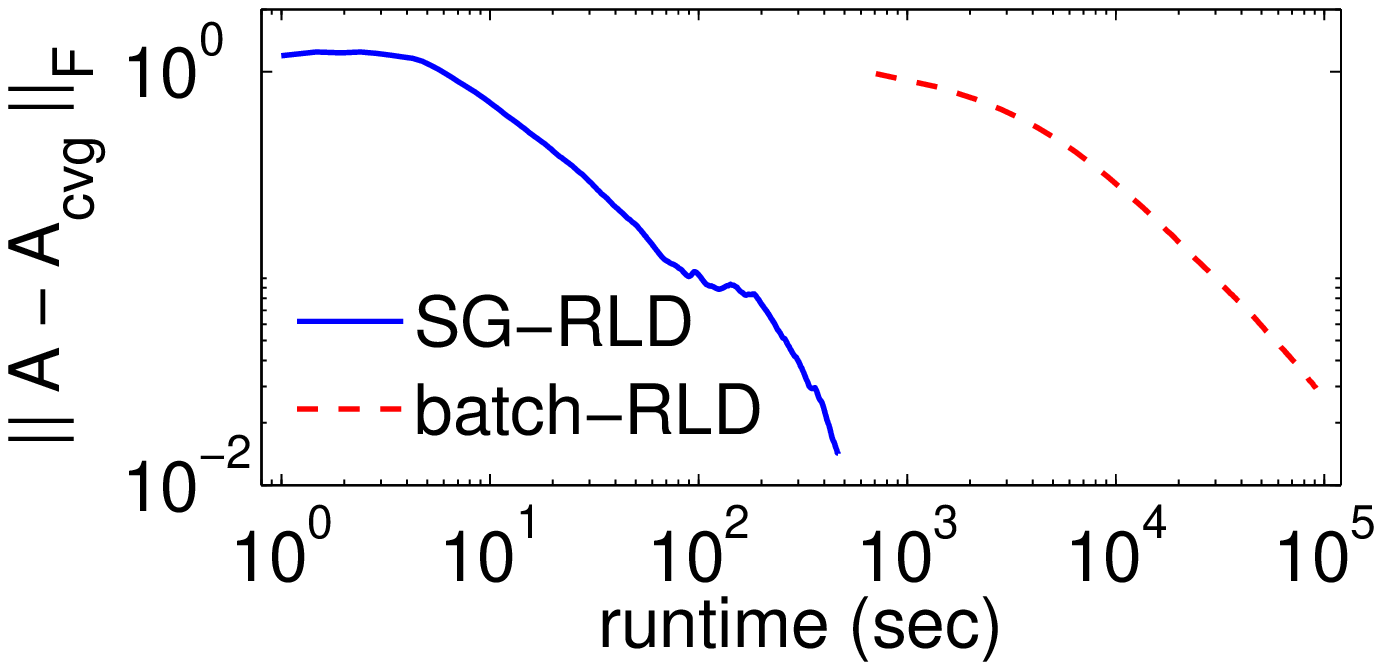}
\vspace{-0.2in}
\caption{\small Inference of ion channel data
\emph{Top:} SG-RLD segmentation at runtimes: 44.05, 138.51, and 466.82 (sec).
\emph{Bottom:} Batch-RLD segmentation at runtimes: 716.19, 2124.43, and 7245.14 (sec).
\emph{Right:} Held-out-probability of $5000$ unobserved data points (top) and error decay of transition matrix estimates (bottom) for SG-RLD and batch-RLD methods in loglog scales.
$A_{\rm cvg}$ denotes the estimated transition parameters $A$ after
    convergence. SG-RLD obtains plausible segmenations and accurate estimates
    of the transition matrix in a fraction of the time as a batch algorithm.
\label{fig:ion}
}
\vspace{-0.05in}
\end{figure*}
\subsection{Non-conjugate Emission Distributions}
We next demonstrate the benefit of our SG-RMCMC algorithm in being able to
handle non-conjugate emissions, an essential feature to perform flexible Bayesian analyses.
We simulate $2\times 10^5$ observations from a $2$-state HMM with log-normal emissions.
Details of the parameter settings used to generate the data are in the
Supplement.
We evaluate the ability of two different HMM models in terms of parameter
estimation and model selection accuracy.
%to both segment and predict unobserved obserations. Fig.~\ref{fig:non-conj} (top) depicts a subset of the synthetic observations.

The first HMM we consider uses log-normal emissions
with non-conjugate normal priors. The second model uses Gaussian emissions with a
conjugate normal-inverse-Wishart prior.
In Fig.~\ref{fig:non-conj} we show that the non-conjugate model obtains accurate estimates of the transition matrix in substantially fewer iterations than the conjugate model.
Next, we demonstrate that efficiently handling non-conjugate models leads to
improved model selection. Specificallly, we use SG-RLD to fit both the
conjugate and non-conjugate HMMs described above with $K=1,2,3,4$ states and compute the marginal likelihood of the observations under each model.
In the table of Fig.~\ref{fig:non-conj} we see that the non-conjugate model
selects the right number of states ($2$), whereas the conjugate model selects a
model with more states ($4$).
The ability to use non-conjugate HMMs for truly massive data sets has been
infeasible until this point and this experiment demonstrates its utility.

\subsection{Ion Channel Recordings}

We investigate the behavior of the SG-RLD sampler
on ion channel recording data.
In particular, we consider a 1MHz recording from~\citet{Rosenstein:2013} of a
single alamethicin channel. This data was previously investigated
in~\citet{Palla:2014} and \citet{Tripuraneni:2015} using a complicated Bayesian
nonparametric HMM. In that work, the authors downsample the data by a factor of
$100$ and only used $10,000$ and $2,000$ observations respectively due to the challenge of scaling computations to the full sequence.
We subsample the time series by a factor of $50$, resulting in $209,634$
observations, to reduce the strong autocorrelations present in the observations
that are not captured well by a vanilla HMM. However, our algorithm would have
no difficulty handling the full dataset. We further log-transform and normalize the observations to use Gaussian emission.

%\begin{figure}[t!]
%\includegraphics[height=0.12\textheight]{pics/segments_1.eps}
%\includegraphics[height=0.12\textheight]{pics/segments_complete_1.eps}\\
%\includegraphics[height=0.12\textheight]{pics/segments_10.eps}
%\includegraphics[height=0.12\textheight]{pics/segments_complete_10.eps}\\
%\includegraphics[height=0.12\textheight]{pics/segments_3.eps}
%\includegraphics[height=0.12\textheight]{pics/segments_complete_3.eps}
%\caption{\small Ion channel recordings
%\emph{Top:} SG-RLD segmentation at runtime: 44.05, 138.51, and 466.82 (s).
%\emph{Bottom:} batch-RLD segmentation at runtime: 716.19, 2124.43, and 7245.14 (s).
%\label{fig:ion}
%}
%\end{figure}

We use a non-informative flat prior to analyze the ion channel data.
In Fig.~\ref{fig:ion} we see that before the batch-RLD algorithm finishes a single iteration, the SG-RLD algorithm has already converged and generated a reasonable segmentation.
With the converged estimation of the transition parameters $A$ as reference, we calculated the speed of convergence of SG-RLD and batch-RLD algorithms and found that the SG-RLD is approximately $1,000$ times faster.

%\begin{figure}[t!]
%\includegraphics[height=0.15\textheight]{pics/error_A.eps}
%\includegraphics[height=0.15\textheight]{pics/error_A_whole.eps}
%\caption{\small Convergence of transition matrix.
%\emph{Left:} Error decay of estimated transition matrix for SG-RLD method.
%\emph{Right:} Error decay of estimated transition matrix for batch-RLD method.
%$A_{\rm cvg}$ denotes the estimated transition parameters $A$ after convergence.
%\label{fig:ion_error}
%}
%\end{figure}

%\begin{figure}
%\centering
%  \begin{minipage}[b]{.33\textwidth}
%    {\label{fig:figA}\includegraphics[width=\textwidth,height=5cm]{figA}}
%  \end{minipage}
%\begin{minipage}[b][s]{.33\textwidth}
%\centering
%  {\label{fig:figB}\includegraphics[width=\textwidth,height=2cm]{figB}}
%\vfill
%  {\label{fig:figC}\includegraphics[width=\textwidth,height=2cm]{figC}}
%\end{minipage}
%\caption{my caption. (a) is .... (b) is .... (c) is ....}
%\label{fig:Test}
%\end{figure}

\section{Discussion}

We have developed an SG-MCMC algorithm to perform inference in
HMMs for massive observation sequences. The algorithm
can be used with non-conjugate emission distributions and is thus applicable to
modeling a variety of data. Also, the algorithm asymptotically samples from the
true posterior as opposed to variational approaches.

Developing the algorithm relied on three ingredients. First, we derived
an efficient approach to estimate the gradient of the marginal likelihood of
the HMM from only small subchains. Second, we developed a principled approach
using buffers to mitigate the errors introduced when breaking the dependencies
at the boundaries of the subchains. Unlike previous heuristic buffering schemes, our
approach is theoretically justified using random dynamical systems. Last, we
utilize sampling scheme based on the mixing time of the HMM to
ensure subchains are approximately independent.

In future work we will extend these ideas to other models of dependent data,
such as Markov random fields. Also, the ideas presented here are not limited to
MCMC and could be used to develop more principled variational inference
algorithms for dependent data.

%TODO Nick: Use full TerraSwarm ack (see email)
\subsection*{Acknowledgements}
This work was supported by ONR Grant N00014-15-1-2380, NSF CAREER Award IIS-1350133, and by TerraSwarm, one of six centers of STARnet, a Semiconductor Research Corporation program sponsored by MARCO and DARPA.
NF was supported by a Washington Research Foundation Innovation Postdoctoral Fellowship in Neuroengineering and Data Science.

\newpage
\onecolumn
\appendix
\section{Gradient of the Posterior}

For the hidden Markov model (HMM), the posterior distribution of all hyperparameters $\theta$ can be calculated by the Bayes rule, where
\[
p(\theta | \vy) \propto p(\vy|\theta) p(\theta).
\]
Since
\[
p(\vy,\vx|\theta) = \pi_0(x_0) \prod_{t=1}^T A_{x_t,x_{t-1}}
\cdot \prod_{t=1}^T p(y_t|x_t),
\]
where $\vy = (y_1, \cdots y_T)$ denotes the data as real valued vector, and $\vx = (x_1, \cdots x_T)$ as discrete valued vector with $x_t \in \{1, \cdots K\}, \forall t$.
We can directly marginalize out the hidden variables, $\vx$, with matrix multiplication as
\[
p(\vy|\theta) = \vone_T^\rt \ P(y_T) A \cdots P(y_1) A \ \vpi_0,
\]
where $P(y_T)$ is a diagonal matrix and $P_{i,j}(y_t) = p(y_t | x_t=i) \delta_{i,j}$;
$\vone_T^\rt = (1,\cdots,1)$ is a row vector of $k$ ones ($ ^\rt$ denotes transpose);
$(\vpi_0)_i = \pi_0(x_0 = i)$.
%vector $(\vpi_0)_i$ denotes initial probability of the hidden variable $x_0$ in state $i$.
Hence the same as Eq.~(2), (3) and (8) of the main paper, the posterior distribution is:
\[
p(\theta|\vy) = \vone_T^\rt \ P(y_T) A \cdots P(y_1) A \ \vpi_0 \cdot p(\theta).
\]

When we divide the whole sequence into subsequences of
\[
\vy_{\tau,L} = (y_{\tau-L},\ldots,y_\tau,\ldots,y_{\tau+L}),
\]
the posterior can be rewritten as:
\begin{align}
	p(\theta|\vy) \propto \vone^\rt \prod_{\vy_{\tau,L} \in \mathcal{S}}P(\vy_{\tau,L}) \vpi_0 \cdot p(\theta)
	\label{eq:hmm_post_short},
\end{align}
where $\mathcal{S}$ is the minimum set of $\vy_{\tau,L}$ covering $\vy$.

We can then use gradient information of the posterior distribution to construct MCMC algorithms.
The gradient of the log-posterior distribution is:
%\[
%\dfrac{\partial \ln p(\theta | \vy)}{\partial \theta_i}
%= \sum_{j=1}^T \dfrac{ \vq_0^\dag P(y_T) A \cdots P(y_{j+1})A
%\dfrac{\partial \left(P(y_j) A\right)}{\partial \theta_i} \cdots P(y_1) A \vpi_0 }
%{ \vq_0^\dag P(y_T) A \cdots P(y_j) A \cdots P(y_1) A \vpi_0 }
%\]
\[
\dfrac{\partial \ln p(\theta | \vy)}{\partial \theta_i}
= \sum_{\tau=1}^{\snorm} \dfrac{ \vone^\rt P(\vy_{\snorm,L}) A \cdots
\dfrac{\partial \left(P(\vy_{\tau,L}) A\right)}{\partial \theta_i} \cdots P(\vy_{1,L}) A \vpi_0 }
{ \vone^\rt P(\vy_{\snorm,L}) A \cdots P(\vy_{\tau,L}) A \cdots P(\vy_{1,L}) A \vpi_0 }
+ \dfrac{\partial \ln p(\theta)}{\partial \theta_i}.
\]
Denote $\vq_{\tau+L+1}^\rt = \vone_T^\rt P(y_T) A \cdots P(y_{t+1}) A$ and $\vpi_{\tau-L-1} = P(y_{t-1}) A \cdots P(y_1) A \vpi_0$. Then
\begin{align}
\label{eq:gradU_full}
\dfrac{\partial U(\theta)}{\partial \theta_i} &= -\dfrac{\partial \ln p(\vy | \theta)}{\partial \theta_i} - \dfrac{\partial p(\theta)}{\partial \theta_i}
\nonumber\\
&= -\sum_{\vy_\tau\in \widetilde{\mathcal{S}}} \dfrac{\vq_{\tau+L+1}^\rt \dfrac{\partial P(\vy_\tau)}{\partial \theta_i} \vpi_{\tau-L-1}}{\vq_{\tau+L+1}^\rt P(\vy_\tau) \vpi_{\tau-L-1}}
- \dfrac{\partial \ln p(\theta)}{\partial \theta_i},
\end{align}
as shown in Eq.~(11) of the main paper.

\section{Lyapunov Exponent}
The question of buffer length is equivalent to: for two random vectors $\vpi$ and $\vpi^*$, what's the expected length of $LB$ such that after the application of $P(\vy_{LB})$, $\vpi$ and $\vpi^*$ will synchronize?
This is a question of random dynamical systems and can be answered through defining the \textit{Lyapunov exponent}.

We first transform $\vpi$ through stereographic projection into $K-1$ dimensions and denote as: $\vr$.
Then operator $P(y_t)A [\ \cdot\ ]$ is projected to new space and the equivalent dynamics over $\vr$ becomes: $F_{y_t}$.
We define the Lyapunov exponent $\mathfrak{L}$ through the projected random dynamics $F_{y_t}$ as
\begin{align}
\mathfrak{L} = \int_{\Omega\times\mathbb{R}^{K-1}}\ln|| \nabla_{\vr} F_{y}(\vr) || \rd \mu_{y} \rd \mu_{\vr},
\end{align}
where $y\in\Omega$.
Measure $\mu_{y}$ corresponds to the distribution of the data $y_t$, and $\mu_{\vr}$ is the invariant measure of $\vr$ under the dynamics of $P(y_t)A$, which will be estimated through sampling.

Once the Lyapunov exponent $\mathfrak{L}$ is calculated, we can set the buffer length:
\begin{align}
    B = \dfrac{1}{\mathfrak{L}} \ln\left(\dfrac{\delta}{\delta_0}\right),
\end{align}
where $\delta=10^{-3}$ is the error tolerance and $\delta_0 = 2$ is the maximum initial error for probability vectors.

\section{Subsequence Sampling Procedure}

We use the following sampling procedure to obtain the subsequences used to compute stochastic gradient estimates. In order to enforce the non-overlapping mixing-time constraint between adjacent subsequences, we sample them sequentially. This results in the following form for the probability of the minibatch $\widetilde{\mathcal{S}}$:
%$p(\widetilde{\mathcal{S}}) = \sum_{n=0}^{R-1} \dfrac{L}{T - n(L + 2B + 2L)}$
$p(\widetilde{\mathcal{S}}) = \prod_{n=0}^{R-1} L/|\mathcal{S}_n|$,
where $|\mathcal{S}_0| = T$, $|\mathcal{S}_n| = |\mathcal{S}_{n-1}| - (\nu + 2B + 2L) - L_{\mathrm{overlap}}$.
The quantity $L_{\mathrm{overlap}}$ is calculated as follows:
\[
L_\mathrm{overlap}^0 = |\tau_n|,
\]
\[
L_\mathrm{overlap}^T = |T - \tau_n|,
\]
\[
L_\mathrm{overlap}^{L} = \min_{n'=1, \tau_{n'}<\tau_n}^{n-1}\{|\tau_n - \tau_{n'}|\} - L - B,
\]
\[
L_\mathrm{overlap}^{R} = \min_{n'=1, \tau_{n'}>\tau_n}^{n-1}\{|\tau_n - \tau_{n'}|\} - L - B.
\]
If $\min\{L_\mathrm{overlap}^0, L_\mathrm{overlap}^T, L_\mathrm{overlap}^{L}, L_\mathrm{overlap}^{R}\} \geq 2\nu+3L+3B$,
the minimum number of observations required to fit an entire subsequence while respecting minimum gap $\nu$,
$L_{\mathrm{overlap}} = 0$.
%if the center of the $n$th subsequence, $\tau_n$, is at least $2\nu+3L+3B$ time indices away from the boundary of adjacent subsequences and if $\tau < 2\nu+3L+3B$ and $T-\tau < 2\nu+3L+3B$.
Otherwise, $L_{\mathrm{overlap}}$ equals to the sum of all the above terms that are less than $2\nu+3L+3B$.

Since $T\gg L, B, \nu$, then $p(\widetilde{\mathcal{S}})$ provides the correct probability of the minibatch $\widetilde{\mathcal{S}}$.

\section{Detailed Descriptions of Experiments}
\subsection{Evaluating Buffer Effectiveness}
\begin{figure}[t!]
%\twocolumn[
%\includegraphics[height=0.23\textwidth]{pics/DD_demo}
%\includegraphics[height=0.23\textwidth]{pics/RC_demo}\\
%\includegraphics[height=0.185\textwidth]{pics/diag_dom_error_A.eps}
%\includegraphics[height=0.185\textwidth]{pics/rev_cyc_error_A.eps}
%\vspace{-0.1in}
\centering{
\includegraphics[height=0.4\textwidth]{DD_demo} \hspace{8pt}
\includegraphics[height=0.4\textwidth]{RC_demo}\\
\includegraphics[height=0.37\textwidth]{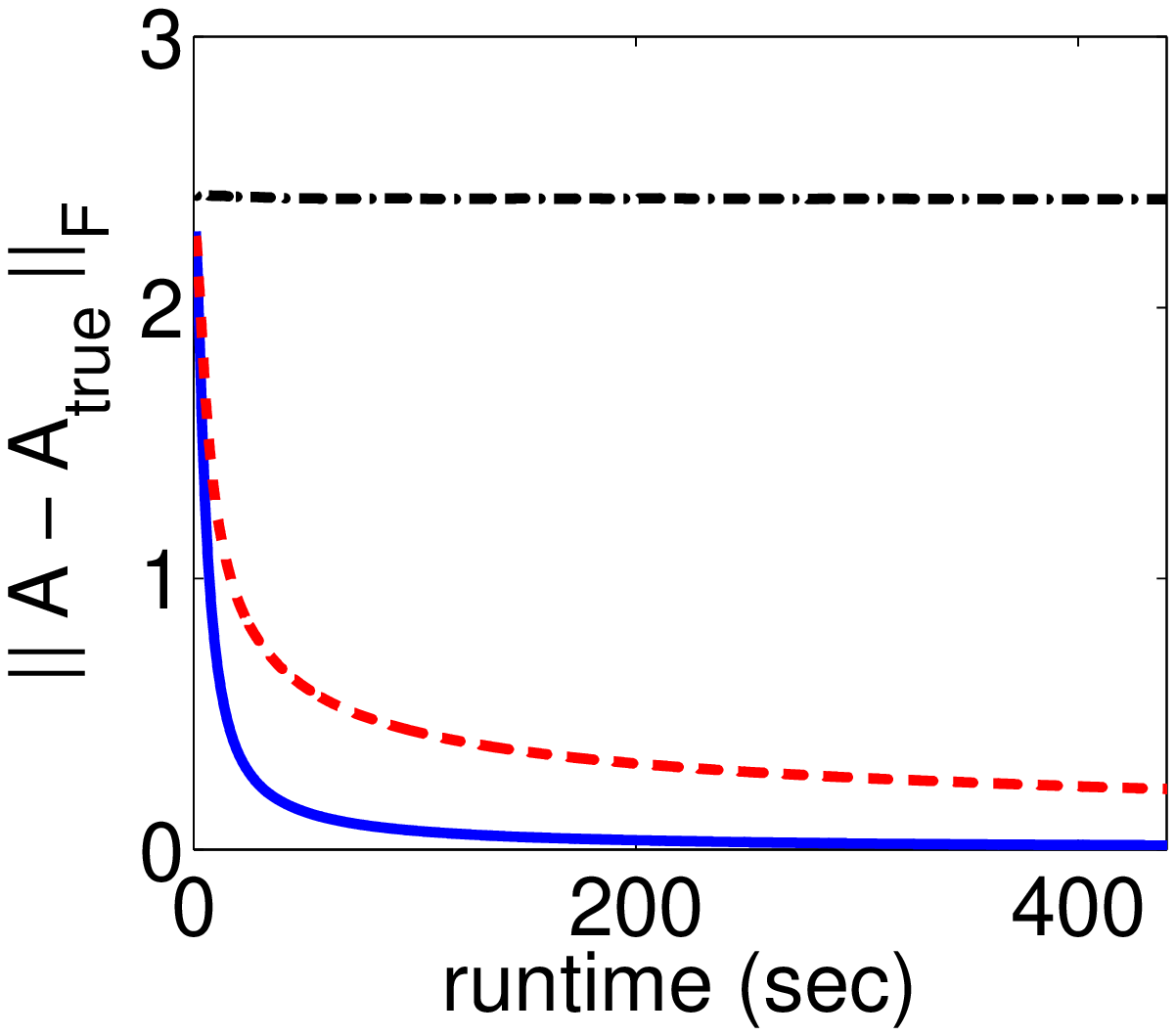}
\includegraphics[height=0.37\textwidth]{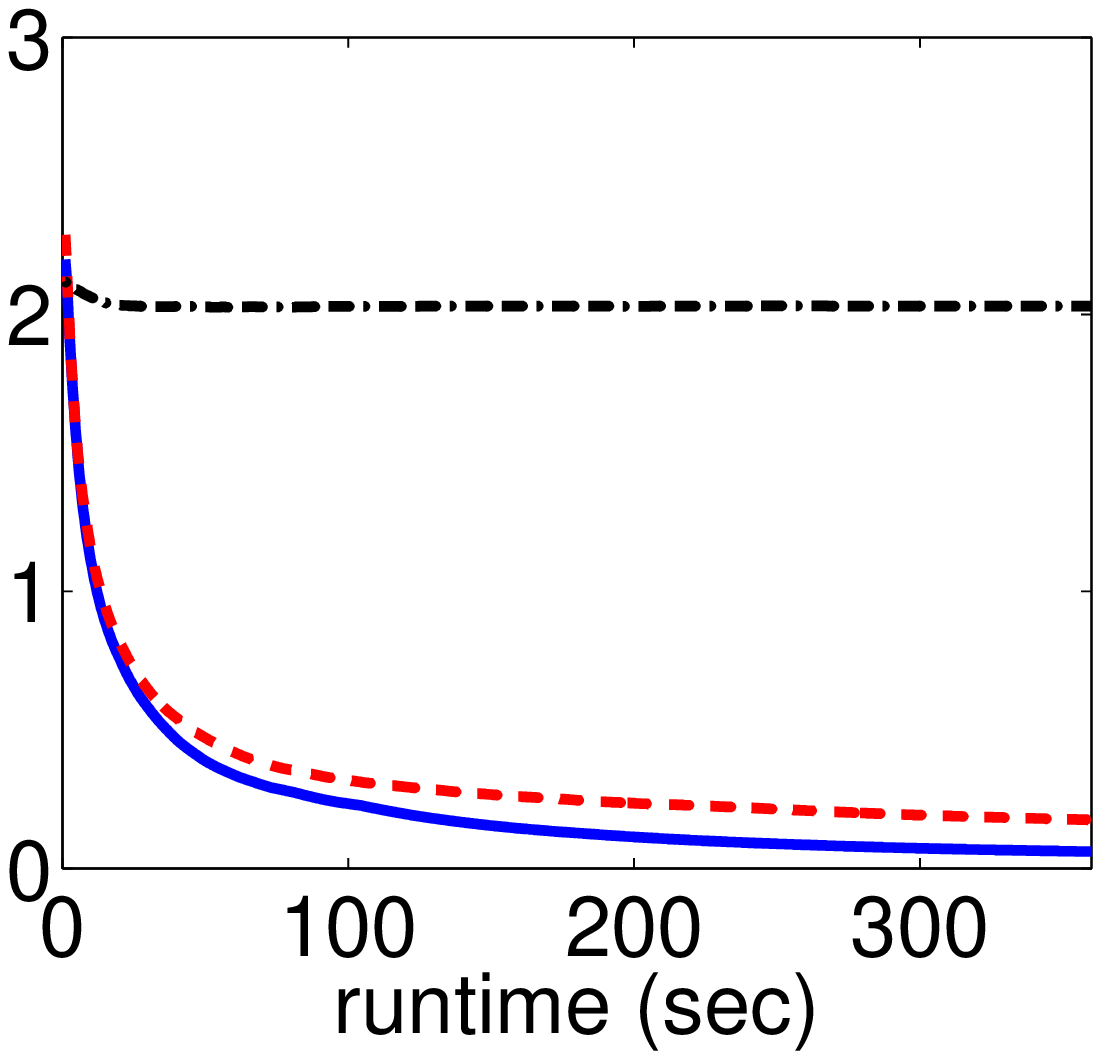}
}
\caption{\small Synthetic experiments with hard-to-capture dynamics.
Diagonally dominant (DD) (\emph{left}) and reversed cycles (RC) (\emph{right}) experiments.
\emph{First Row:} The emission distributions corresponding to $8$ different states.
Arrows in the RC case indicate the Markov transition structure with transition between bridge states as dashed arrows.
\emph{Second Row:} Decrease of error in transition matrix estimation versus runtime.
Comparisons are made for SG-RLD algorithms with estimated buffer, without buffer, and treating data as i.i.d.
All of the experiments use a constant computation budget by varying the number of subchains, $|\tilde{S}|$, with the length of the subchains, $L$.
}\vspace{-0.1in}
%]
\label{fig:sim_appnd}
\end{figure}
The first data set, \textit{diagonally dominant} (DD) consists of a Markov chain that heavily self-transitions. Most subchains in a minibatch thus contain redundant information with observations generated from the same latent
state. Although transitions are rarely observed, the emission means are set to be distinct so that this
example is likelihood-dominated and highly identifiable. See Fig.~\ref{fig:sim_appnd} (top left).
%For this data we set the subchain length $L=2$ and use $|\widetilde{S}| = 10$
For this data we choose $L=2$ and $|\widetilde{S}|=10$ subsequences in order to incorporate observations from distant parts of the observation sequence.
%minibatches since the large number of self-transitions makes the information in long subchains redundant.
This corresponds to an extreme setting where each gradient is based only on $5$ observations. The transition matrix and emission parameters used for this experiment were:

\[ A_{DD} = \left( \begin{array}{cccccccc}
.999 & .001 & 0 & 0 & 0 & 0 & 0 & 0 \\
0 & .999 & .001 & 0 & 0 & 0 & 0 & 0 \\
0 & 0 & .999 & .001 & 0 & 0 & 0 & 0  \\
0 & 0 & 0 & .999 & .001 & 0 & 0 & 0 \\
0 & 0 & 0 & 0 & .999 & .001 & 0 & 0 \\
0 & 0 & 0 & 0 & 0 & .999 & .001 & 0 \\
0 & 0 & 0 & 0 & 0 & 0 & .999 & .001 \\
.001 & 0 & 0 & 0 & 0 & 0 & 0 & .999
\end{array} \right) . \]

\[\boldsymbol{\mu}_{DD} = \left\{
(0,20); (20,0); (-30,-30); (30,-30); (-20,0); (0,-20); (30,30); (-30,30);
\right\}\]
and $\Sigma_{DD} = I$ for all states.

The second dataset we consider contains two \textit{reversed cycles} (RC): the Markov chain strongly transitions
from states $1 \rightarrow 2 \rightarrow 3 \rightarrow 1$ and $ 5 \rightarrow 7 \rightarrow 6 \rightarrow 5$ with a small probability of transiting
between cycles via bridge states $4$ and $8$. See Fig.~\ref{fig:sim_appnd} (top
right). The emission means for the two cycles are very similar
but occur in reverse order with respect to the transitions.
The emission variance is larger, making states $1$ and $5$, $2$ and $6$, $3$ and $7$ indiscernible by themselves.
Transition information in observing long
enough dynamics is thus crucial to identify between states $1, 2, 3$ and $5, 6,
7$.
Therefore, we set $L=5$ and $|\widetilde{S}|=4$.
Note that same amount of data are used in the calculation of the gradient.
The transition matrix and emission parameters were:

\[ A_{RC} = \left( \begin{array}{cccccccc}
.01 & 0 & .85 & 0 & 0 & 0 & 0 & 1 \\
.99 & .01 & 0 & 0 & 0 & 0 & 0 & 0  \\
0  & .99 & 0 & 0 & 0 & 0 & 0 & 0 \\
0 & 0 & .15 & 0 & 0 & 0 & 0 & 0 \\
0 & 0 & 0 & 1 & .01 & 0 & .85  & 0 \\
0 & 0 & 0 & 0 & .99 & .01 & 0 & 0 \\
0 & 0 & 0 & 0 & 0 & .99 & 0 & 0 \\
0 & 0 & 0 & 0 & 0 & 0 & .15 & 0
\end{array}\right) . \]

\begin{equation*}
\boldsymbol{\mu} = \left\{ (-50,0); (30,-30); (30,30); (-100,-10); (40,-40); (-65,0); (40, 40); (100,10)  \right\},
\end{equation*}
and $\Sigma_{RC} = 20*I$ for all states.

We use a non-conjugate flat prior to demonstrate the flexibility of our
algorithm.
We initialize with a short run of k-means clustering to ensure that different states have different emission parameters.

\subsection{Non-conjugate Emission Distribution}
For the non-conjugate experiment, we used the following transition matrix:
\[
\left( \begin{array}{cc}
.1 & .9 \\
.9 & .1
\end{array}\right).
\]
For emission probability, we use a log-normal distribution: $p_k(y) \propto e^{-\dfrac{\ln (y-\mu_k)^2}{2\sigma_k^2}}$
with parameters: $\mu_1=0$, $\mu_2=4$; $\sigma_1=\sigma_2=2$.

In the non-conjugate model, we use the following priors on the emission parameters:
$\mu_1, \mu_2, \sigma_1, \sigma_2 \sim \mathcal{N}(0,1)$.

\end{document}